\documentclass[remotesensing,article,accept,pdftex,moreauthors]{Definitions/mdpi} 

\firstpage{1} 
\pubvolume{1}
\issuenum{1}
\articlenumber{0}
\pubyear{2025}
\copyrightyear{2025}
\externaleditor{Firstname Lastname}
\datereceived{6 December 2024} 
\daterevised{9 January 2025} 
\dateaccepted{10 January 2025} 
\datepublished{ } 
\hreflink{https://doi.org/} 

\usepackage{algorithm,algcompatible}
\DeclareMathOperator*{\argmin}{argmin}
\algnewcommand\INPUT{\item[\textbf{Input:}]}%
\algnewcommand\OUTPUT{\item[\textbf{Output:}]}%
\usepackage{graphicx}
\usepackage{textcomp}
\usepackage{xcolor}
\usepackage{color}
\usepackage{diagbox}
\usepackage{soul}
\usepackage[normalem]{ulem}
\Title{Self-Supervised Deep Hyperspectral Inpainting with Plug-and-Play and Deep Image Prior Models} 

\TitleCitation{Self-Supervised Deep Hyperspectral Inpainting with Plug-and-Play and Deep Image Prior Models}

\Author{{Shuo} 
 Li $^{1,*}$ and Mehrdad Yaghoobi $^{2}$}

\AuthorNames{Shuo Li and Mehrdad Yaghoobi}

\AuthorCitation{Li, S.; Yaghoobi, M.}

\address{%
$^{1}$ \quad {School of Engineering, University of Edinburgh, United Kingdom, EH8 9YL; s1809498@ed.ac.uk}  
 \\
$^{2}$ \quad {Institute of Imaging, Data and Communications, University of Edinburgh, United Kingdom, EH8 9YL; m.yaghoobi-vaighan@ed.ac.uk}
}

\corres{Correspondence: s1809498@ed.ac.uk}

\abstract{Hyperspectral images are typically composed of hundreds of narrow and contiguous spectral bands, each containing information regarding the material composition of the imaged scene. However, these images can be affected by various sources of noise, distortions, or data loss, which can significantly degrade their quality and usefulness. This paper introduces a convergent guaranteed algorithm, LRS-PnP-DIP(1-Lip), which successfully addresses the instability issue of DHP that has been reported before. The proposed algorithm extends the successful joint low-rank and sparse model to further exploit the underlying data structures beyond the conventional and sometimes restrictive unions of subspace models. A stability analysis guarantees the convergence of the proposed algorithm under mild assumptions , which is crucial for its application in real-world scenarios. Extensive experiments demonstrate that the proposed solution consistently delivers visually and quantitatively superior inpainting results, establishing state-of-the-art performance. 
}


\begin{document}

\section{Introduction}
\label{sec:introduction}
Hyperspectral remote sensing {has found widespread application} in numerous fields such as astronomy, agriculture, environmental monitoring, and Earth observation. Hyperspectral images (HSIs) are often captured via satellite or airborne sensors, each presenting samples at different time slots, using push-broom strategies along the flying \mbox{pathway \cite{ortega2019hyperspectral}}. The nature of the HSI acquisition system makes the HSI a high-resolution 3D data cube that covers hundreds or thousands of narrow spectral bands, conveying a wealth of spatio-spectral information. However, due to instrumental errors, imperfect navigation and atmospheric {variations}, practical HSIs can suffer from noise, and missing pixels or even entire lines of pixels ({an example of this type of missing data can be found in the recently released Earth Surface Mineral Dust Source Investigation (EMIT) hyperspectral dataset, developed by NASA’s Jet Propulsion Laboratory and launched on 14 July 2022 \cite{EMIT}}). These issues can severely impact {the accuracy and reliability of subsequent analyses}, making HSI inpainting a critical task in the field of remote sensing and Earth observation. 

\subsection{Hyperspectral Image Inpainting}
Hyperspectral image inpainting {refers to} the restoration of missing or corrupted data in acquired HS images. Unlike RGB images, for which inpainting involves filling a single pixel value (red, green, and blue channels), HSI inpainting requires filling in a complex vector that contains extensive spectral information. This additional complexity makes the already challenging task of HSI inpainting even more difficult. The primary objective of HSI inpainting is to create visually convincing structures while ensuring that the texture of the missing regions is spectrally coherent. The inpainting of HS images typically involves two main steps: (a) estimating the missing or corrupted data by leveraging information from the neighboring spectral bands and spatially adjacent pixels and (b) refining the estimated data to ensure consistency with the overall spectral and spatial properties of the HS image.
Traditional approaches, such as those proposed by \cite{zhuang2018fast}, assume that the spectral vectors of the HSI always exist in some unknown low-dimensional subspaces, and the missing regions are estimated through projections. {However, the performance of such techniques deteriorates significantly when large portions of the image are missing}, and they may even fail when all spectral bands are absent. Another line of research involves the tensor completion process to restore incomplete observations \cite{zhao2021tensor,li2021tensor,luo2022self}. {These methods leverage the spatio-spectral correlations of the hyperspectral data cube, offering potential improvements over subspace-based techniques.} \par
{Over} the past decade, the success of deep learning has brought new opportunities for solving HSI inpainting tasks. {In \cite{cite_attention_cloud_transformer, cite_attention_cloud_transformer_2}, the authors proposed transformer-based frameworks to effectively capture both local and global contextual information in 3D images, demonstrating the potential of attention modules in exploring relationships between missing pixels and background pixels.} In \cite{wong2020hsi}, the authors showed that missing pixels can be predicted using generative neural networks. {However, this approach has two key limitations:} (1) it requires training the networks on extensive datasets to achieve the desired performance, and (2) the inpainted areas often suffer from over-smoothing, resulting in the loss of critical information, such as {abrupt} changes in the surface materials. More recently, researchers discovered that the network's structure itself may serve as a good inductive bias for regularizing image reconstruction tasks, such as image inpainting \cite{ulyanov2018deep}. A deep image prior (DIP) enables the learning of priors directly from the image without the need for extensive training datasets, which was further developed and successfully applied to hyperspectral (HS) images \cite{sidorov2019deep}. Following \cite{sidorov2019deep}, subsequent works \cite{2021_DIP_In_Loop,lai2022deep, wu2022adaptive, Ulugbek_Learned_Priors, niresi2023robust,hong2023decoupled_self_supevised,added_ref_1} have shown that certain trained or untrained neural networks can be directly incorporated into iterative solvers to improve the reconstruction accuracy, i.e., as the regularizer.
{More recently, the authors of \cite{DDS2M} introduced a self-supervised framework that leverages the powerful learning capabilities of diffusion models and sets new state-of-the-art benchmarks across several HSI reconstruction tasks.} {However, as it was reported in our preliminary report \cite{li2023self}, as well as in a series of DIP-related works \cite{Deep_Decoder,2019_Deep_Red,2019_DIP_TV,lai2022deep}, the DIP may not always be the ideal solution, despite promising empirical results. On the practical side, this is because a DIP works by directly fitting a neural network to the corrupted HS image; in the absence of regularization or explicit denoising mechanisms, a DIP tends to overfit, which will eventually capture the noise in the image. On the theoretical side, a DIP lacks a strong theoretical foundations to explain its performance, and its convergence is highly sensitive to random initialization of the neural network. Different initializations could potentially lead to poor results. Although subsequent works \cite{pnp_dip,2021_DIP_In_Loop,lai2022deep} have attempted to mitigate this issue by introducing a plug-and-play (PnP) denoiser into the training of a DIP, they require either a suitable early stopping criterion or the manual control of the learning rate. As a result, there remains a need for a new HSI inpainting algorithm that combines the strengths of both PnP and DIP methods while offering convergence guarantees.}

\subsection{Sparsity and Low Rankness in HSIs}
{Despite the high dimensionality, HS images exhibit intrinsic} sparse and low-rank structures due to the high correlation between spatial pixels and spectral channels. This property has made sparse representation- (SR) and low rank- (LR) based methods popular for the processing of HSIs \cite{sparse_representation,low_rank,li2023lrr_low_rank_net}. SR relies on the key assumption that the spectral signatures of the pixels approximately lie in a low-dimensional subspace spanned by representative pixels from the same class. With a given dictionary, SR allows for sparse decomposition of HSIs into a linear combination of several atoms, thus exploiting the spatial similarity of HSIs ({It is worth mentioning that, when the dictionary is not given, we can manually use the end-members of some pixels with pure spectra and build the dictionary, or learn the dictionary \cite{dictionary_learning} in an unsupervised manner, using sparsity as a regularizer}). {This assumption about the underlying structures} has been effectively applied in various HSI tasks, including HSI classification \cite{SR-application_1}, denoising \cite{LR_SR_1}, and unmixing \cite{SR-application_3}. \par
{LR} is another widely used regularizer in HSI processing, which assumes that pixels in the clean HSI image have high correlations in the spectral domain, thus capturing both the spectral similarity and global structure of HSIs. {LR allows for the modeling of the HSI as a low-rank matrix; thus, the missing and corrupted information in the HSIs can be effectively reconstructed by leveraging the underlying low-rank structure. In many state-of-the-art HSI reconstruction algorithms, LR is often used in parallel with SR to achieve better reconstruction accuracy \cite{LR_SR_1,LR_SR_2}. By combining the two priors together, algorithms can better handle missing data, noise, and corruption while maintaining the overall structure and spectral integrity of the image.} For more detailed information on LR and SR of hyperspectral images, we refer readers to the review work in \cite{SR_LR_review}.  \par
{However}, a notable limitation with such an LR or SR model is that the real HS data do not exactly follow the sparse and/or low-dimension assumptions.; e.g., there are long trails (numerous small but non-zero elements) in the descending sorted absolute sparse coefficients and singular values, where overly-restrictive assumptions on these coefficients could lead to under-fitting. {In conclusion, both LR and SR rely on some predefined mathematical models that assume certain properties of the HS image, such as low-dimensional manifolds (low-rankness) or sparse representations in specific domains. While these models are useful, they struggle to capture more complex patterns in the real HS data, such as non-linear relationships and complex spatio-spectral dependencies. Moreover, when a significant percentage of the data are corrupted or entire spectral bands of pixels are missing, these methods often fail \cite{zhuang2018fast}. The question of how deep neural models can address this issue by exploiting more data-adaptive, low-dimensional models in a self-supervised learning framework is the main topic of this study. } \par
{In} this paper, we start by giving a brief overview of the LRS-PnP-DIP algorithm \cite{li2023self}; we replaced the denoising/thresholding step with the PnP denoiser, and we used a DIP to replace the low-rank component of the formulation. Since there remains a lack of comprehensive theoretical analysis of the DIP and its extensions that can guarantee the convergence, we demonstrate both theoretically and empirically that small modifications can ensure the convergence of LRS-PnP and LRS-PnP-DIP, while the original algorithms in \cite{li2023self} may diverge. We propose a variant of the LRS-PnP-DIP algorithm dubbed LRS-PnP-DIP(1-Lip), which not only outperforms existing learning-based methods but also resolves the instability issue of a DIP in an iterative algorithm. 

\subsection{Contributions}
This paper presents a novel HS inpainting algorithm that leverages the learning capability of deep networks {while providing} insights into the convergence behavior of such complex algorithms. {Through rigorous mathematical analysis, we derive sufficient conditions for stability and fixed-point convergence} and show that these conditions are satisfied by slightly modified algorithms. Our approach offers several key contributions, including the following:
\begin{itemize} 
\item This paper explores the LRS-PnP and LRS-PnP-DIP algorithm in more depth by showing the following: (1) The sparsity and low-rank constraints are both important to the success of the methods in \cite{li2023self}. (2) DIP can better explore the low-rank subspace compared to conventional subspace models such as singular value \mbox{thresholding (SVT}).
\item Under some mild assumptions, a fixed-point convergence proof is provided for the LRS-PnP-DIP algorithm (see Theorem \ref{theorem 1}). We introduce a variant to the LRS-PnP-DIP called LRS-PnP-DIP(1-Lip), which effectively resolves the instability issue of the algorithm by slightly modifying both the DIP and PnP denoiser.
\item To the best of our knowledge, this is the first time the theoretical convergence of PnP with DIP replaced the low-rank prior being analyzed under an iterative framework.
\item Extensive experiments were conducted on real-world data to validate the effectiveness of the enhanced LRS-PnP-DIP(1-Lip) algorithm with a convergence certificate. {The results demonstrate the superiority of the proposed solution over existing learning-based methods in both stability and performance.} 
\end{itemize}\par
{
The remainder of this paper is organized as follows: {Section} \ref{sec2} provides an introduction to our approach, and Section \ref{sec3} describes our main convergence theory for the proposed LRS-PnP-DIP(1-Lip) algorithm. Section \ref{sec4} provides the implementation details and discusses the experimental results. Section \ref{sec5} concludes the paper with a brief discussion of potential future directions.}

\section{Proposed Methods}\label{sec2}
\subsection{Mathematical Formulations}
The task of HSI inpainting can be {formulated} as reconstructing the clean image, $X$, from a noisy and incomplete measurement, $Y$, where an additive noise, $N$, and a masking operator, $M$, are {present:}
\begin{equation} \label{eqn:problem setting}
 Y = {\mathcal{M}}\{X\} +N
\end{equation}\par
{The} clean image $X \in\mathbb{R}^{q}$ (where $q =n_{r} \times n_{c} \times n_{b}$), with $n_{r}$ and $n_{c}$ representing the spatial dimensions of the image, and $n_{b}$ represents the total number of spectral bands. The operator {$ \mathcal{M} \in \mathbb{R}^{q \times q}$} is a binary mask, where zero represents the missing pixels, and one represents the observed and valid pixels. Thus, the masking operator {$\mathcal{M}$} can be represented as a diagonal square matrix. $N$ is the additive Gaussian noise of appropriate size. Usually, $M$ is provided, and the formulation \eqref{eqn:problem setting} {describes a linear system, which can be expressed {as follows:} 
}
\begin{equation} \label{eqn:problem setting_2}
 \boldsymbol{y} = \mathrm{M} \boldsymbol{x} +\boldsymbol{n}
\end{equation}\par
{In} this equation, $\boldsymbol{x}$, $\boldsymbol{y}$, and $\boldsymbol{n}$ represent the vectorized forms of $X$, $Y$, and $N$, respectively, and $\mathrm{M}$ is a diagonal matrix.
We obtain the recovered, i.e., inpainted, image, $\boldsymbol{x^*}$, by applying sparse representation to each image patch, $P_i(\boldsymbol{x})$. {Here, the operator $P_i(\cdot)$ extracts the i-th patch from image $\boldsymbol{x}$, where each patch may consist of only valid pixels or a combination of valid and missing pixels,} depending on the size of $P_i(\cdot)$. The inpainted image $\boldsymbol{x^*}$ can be obtained by solving the following optimization problem:
\begin{equation}
\begin{aligned} \label{eqn:problem setting_3} 
 (\boldsymbol{x^*},\boldsymbol{\alpha^*}) = \argmin_{\boldsymbol{x},\boldsymbol{\alpha}}   \gamma\Vert \boldsymbol{y} -\mathrm{M}\boldsymbol{x} \Vert_{2}^2 + w_{lr}\Vert \boldsymbol{x} \Vert_* \\
 + {\Vert \sum_i(P_i(\boldsymbol{x})-\Phi \boldsymbol{\alpha}_i) \Vert_{2}^2 + w_{s}\sum_i \Vert \boldsymbol{\alpha}_i \Vert_1}\\
\end{aligned}
\end{equation}\par
{The} objective in the expression above {consists of} three terms. The first term is the data fidelity term, weighted by the parameter $\gamma$. Since estimating $\boldsymbol{x}$ from $\boldsymbol{y}$ is inherently ill-posed, i.e., with more unknown variables than equations, the solution is non-unique. {To address this,} we introduce two additional ``priors'' to regularize the inpainting problem: low-rank and sparsity constraints. The second term penalizes solution $\boldsymbol{x}$ to be low-rank, which is typically employed as a surrogate for rank minimization. The third term restricts the missing pixels to be generated from the subspace approximated by the valid pixels. {These terms are weighted by} parameters $w_{lr}$ and $w_{s}$, respectively. The sparse representation problem is solved using a given dictionary, $\Phi$. Specifically, $\Phi$ can be constructed either by using the end-members of some pixels with pure spectra or by learning from the noisy pixels in the observations. It is here exclusively learned from noisy pixels in the observations using online dictionary learning \cite{dictionary_learning}.
{By adopting the augmented Lagrangian and introducing the auxiliary variable $\boldsymbol{u}$ \cite{ADMM}, problem \eqref{eqn:problem setting_3}} can be rewritten as follows:
\begin{equation} \label{opmization_problem}
\begin{aligned}
(\boldsymbol{x^*},\boldsymbol{\alpha^*}) = &\argmin_{\boldsymbol{x}, \boldsymbol{\alpha}} \gamma \Vert \boldsymbol{y} -\mathrm{M}\boldsymbol{x} \Vert_{2}^2 + w_{lr}\Vert \boldsymbol{u} \Vert_*
 + w_{s}\sum_i \Vert \boldsymbol{\alpha}_i \Vert_1 \\
 &+\frac{\boldsymbol{\mu}_1}{2}\Vert \sum_i(P_i(\boldsymbol{x})-\Phi \boldsymbol{\alpha}_i) +\frac{\boldsymbol{\lambda}_1}{\boldsymbol{\mu}_1} \Vert_{2}^2\\
&\textrm{s.t.} \quad \boldsymbol{x} = \boldsymbol{u}
\end{aligned}
\end{equation}\par
{The} Lagrangian multiplier and penalty terms are denoted as $\boldsymbol{\lambda}_1$ and $\boldsymbol{\mu}_1$, respectively. Using the alternating direction method of multipliers (ADMM), we can sequentially update the three variables $\boldsymbol{\alpha}$, $\boldsymbol{u}$, and $\boldsymbol{x}$ to solve problem \eqref{opmization_problem}: \par
{(1)} \textit{Fixing $\boldsymbol{u}$ and $\boldsymbol{x}$, and updating $\boldsymbol{\alpha}$}:
\begin{equation} \label{alpha}
\begin{aligned}
 {\boldsymbol{\alpha}}^{k+1} = \argmin_{\boldsymbol{\alpha}} \frac{\boldsymbol{\mu}_1^k}{2} \sum_i\Vert(P_i(\boldsymbol{x}^k) + \frac{\boldsymbol{\lambda}_1^k}{\boldsymbol{\mu}_1^k})-\Phi \boldsymbol{\alpha}_i \Vert_{2}^2 \\+  w_{s}\sum_i \Vert \boldsymbol{\alpha}_i \Vert_1 \\
\end{aligned}
\end{equation} \par
{The} problem in Equation \eqref{alpha} is a patch-based, sparse coding problem that can be solved using iterative solvers. Let us denote the first term on the right-hand side of Equation \eqref{alpha} as $f(\boldsymbol{\alpha})$ and the second term as $g(\boldsymbol{\alpha})$ while dropping the subscript for simplicity. Problem \eqref{alpha} can be represented as a constrained optimization problem: \\
\begin{equation} \label{alpha_with_f_g_v_substitute}
\begin{aligned}
 (\hat{\boldsymbol{\alpha}},& \hat{\boldsymbol{v}}) = \argmin_{\boldsymbol{\alpha},\boldsymbol{v}} f(\boldsymbol{\alpha}) + w_{s}g(\boldsymbol{v}) \\
 &\textrm{s.t.} \quad \boldsymbol{\alpha} = \boldsymbol{v}
\end{aligned}
\end{equation} \par
{By} introducing an auxiliary variable, $\boldsymbol{u}$, and a multiplier, $\rho$, the augmented Lagrangian of \mbox{\eqref{alpha_with_f_g_v_substitute} becomes the following:}
\begin{equation} \label{ADMM_L_function}
\begin{aligned}
\mathcal{L}(\hat{\boldsymbol{\alpha}}, \hat{\boldsymbol{v}}, \hat{\boldsymbol{u}}) = f(\boldsymbol{\alpha}) + w_{s} g(\boldsymbol{v}) + \boldsymbol{u}^T(\boldsymbol{\alpha}-\boldsymbol{v}) +  \frac{\rho}{2} \Vert \boldsymbol{\alpha}-\boldsymbol{v}\Vert_2^2
\end{aligned}
\end{equation}  \par 
{Then}, ADMM finds solutions by breaking the constrained optimization problem \eqref{alpha_with_f_g_v_substitute} into several sub-problems and updating them separately in an iterative fashion. {The pseudocode is provided in Algorithm \ref{ADMM_pseudo}.} \par

{The} $\boldsymbol{v}^{k+1}$ sub-problem can be solved by the proximal operator \cite{proximal_operator}. The authors in \cite{PnP-ISTA} realized that $\boldsymbol{\alpha}^{k+1}+\boldsymbol{u}^{k}$ can be treated as a ``noisy'' version of $\boldsymbol{v}$, and $g(\boldsymbol{v})$ can be seen as a regularization term. Then, the whole process resembles a denoising operation applying to the intermediate results, which can be simply replaced with the PnP denoisers such as BM3D and the non-local mean (NLM) denoiser. In our approach, we employed PnP-ISTA \cite{PnP-ISTA} as an alternative solution for solving problem \eqref{alpha}. We denote the gradient of $f$ as $\nabla f$ and $I$ as an Identity matrix with an appropriate shape. The entire process can then be replaced with an off-the-shelf denoiser $\mathcal{D}$, operating on $\nabla f$ \cite{PnP-ISTA}. Each iteration takes the following form: \\
\begin{equation}
\begin{aligned} \label{pnp_ista}
 {\boldsymbol{\alpha}^{k+1}} = \mathcal{D}(\mathrm{I}-\nabla f)(\boldsymbol{\alpha}^{k})
\end{aligned}
\end{equation} \par
{PnP-ISTA} can significantly improve the reconstruction quality over traditional ISTA \cite{ISTA} due to the following reasons: (1) it {enables the integration} of state-of-the-art denoisers to handle different noise patterns and different types of images, which is particularly flexible, as one can apply any off-the-shelf denoiser without massively changing the existing framework; and (2) it {allows the use of} denoisers tailored to different problems (e.g., a denoiser that promotes sparsity or smoothness, or even a denoising neural network that encodes more \mbox{complex priors}).

\begin{algorithm}[H]
    \caption{Pseudocode of ADMM for solving problem \eqref{alpha_with_f_g_v_substitute}.} \label{ADMM_pseudo}
    \begin{algorithmic}
     \STATE \textbf{Initialization} $ w_{s}, \rho, k \leftarrow 0, \boldsymbol{\alpha}^{k} \leftarrow      0, \boldsymbol{u}^{k} \leftarrow 0, \boldsymbol{v}^{k} \leftarrow 0$.
         \WHILE{Not Converged}
          \STATE $\boldsymbol{\alpha}^{k+1} \leftarrow \argmin_{\boldsymbol{\alpha}} f(\boldsymbol{x}) + \frac{\rho}{2} \Vert \boldsymbol{\alpha}-(\boldsymbol{v}^{k}-\boldsymbol{u}^{k})\Vert_2^2$.
          \STATE $\boldsymbol{v}^{k+1} \leftarrow \argmin_{\boldsymbol{v}} w_{s} g(\boldsymbol{v}) + \frac{\rho}{2} \Vert \boldsymbol{v}-(\boldsymbol{\alpha}^{k+1}+\boldsymbol{u}^{k})\Vert_2^2$.
          \STATE $\boldsymbol{u}^{k+1} \leftarrow \frac{1}{\rho}\boldsymbol{u}^{k} + ( \boldsymbol{\alpha}^{k+1}- \boldsymbol{v}^{k+1})$.
         \ENDWHILE
    \end{algorithmic}
\end{algorithm} \par
{(2)} \textit{{We} can update $\boldsymbol{u}$ in this setting by}:
\begin{equation}
\begin{aligned}
  {\boldsymbol{u}^{k+1}} = \argmin_{\boldsymbol{u}} w_{lr}\Vert \boldsymbol{u} \Vert_* + \frac{\boldsymbol{\mu_2}^{k}}{2}\Vert (\boldsymbol{x}^k+\frac{\boldsymbol{\lambda_2}^k}{\boldsymbol{\mu}_2^k}) - \boldsymbol{u} \Vert_{2}^2
\end{aligned}
\end{equation}
{where $\boldsymbol{\lambda}_2$ and $\boldsymbol{\mu}_2$ are the corresponding Lagrangian multiplier and penalty term, respectively, which we use subscripts to distinguish between $\boldsymbol{\lambda}_1$ and $\boldsymbol{\mu}_1$ in step \eqref{opmization_problem}.} The problem can be solved using the singular value thresholding (SVT) algorithm \cite{cai2010singular} as follows:
\begin{equation}
\begin{aligned}
  {\boldsymbol{u}^{k+1}} = \mathcal{SVT}(\boldsymbol{x}^k+\frac{\boldsymbol{\lambda}_2^k}{\boldsymbol{\mu}_2^k})
\end{aligned}
\end{equation}\par
{Specifically}, denote $\frac{w_{lr}}{\boldsymbol{\mu}_2^k}$ as $\tau$, matrix $\boldsymbol{x}^k + \frac{\boldsymbol{\lambda}_2^k}{\boldsymbol{\mu}_2^k}$ as $\rm A$, and the soft thresholding operator $\mathcal{S}_\tau$ as follows: 
\begin{equation} \label{SVT_oerator}
\begin{aligned}
  {\mathcal{S}_\tau}(x) = \mathit{sign}(x) \cdot \mathit{max}(\vert x - \tau \vert,0)
  \end{aligned}
\end{equation}
where $sign(\cdot)$ is an operator taking the positive part of its input. Applying \eqref{SVT_oerator} to the singular values of $\rm A$ promotes low-rankness:
\begin{equation} \label{SVT_apply_soft_thresholding}
\begin{aligned}
   {\boldsymbol{u}^{k+1}} = \mathcal{SVT}({\rm A}) = {\rm U_A}\mathcal{S}_{\tau}({\Sigma_{\rm A}}) \rm V_A^T
\end{aligned}
\end{equation}\par
{In} the proposed LRS-PnP-DIP algorithm \cite{li2023self}, this step is substituted with a deep image prior (DIP) $f_\theta(\boldsymbol{z})$, where $\theta$ denotes the network parameters that need to be updated, and the input $\boldsymbol{z}$ is set to $\boldsymbol{x}^k + \frac{\boldsymbol{\lambda}_2^k}{\boldsymbol{\mu}_2^k}$; i.e., the latent image from the previous iterations:
\begin{equation}
\begin{aligned}\label{DIP_replaced}
 {\boldsymbol{u}^{k+1}} = f_\theta(\boldsymbol{x}^k + \frac{\boldsymbol{\lambda}_2^k}{\boldsymbol{\mu}_2^k})
\end{aligned}
\end{equation} \par
{(3)} \textit{{Fixing} $\alpha$ and $\boldsymbol{u}$, and updating $\boldsymbol{x}$}: 
\vspace{-4pt}
\begin{equation}
\begin{aligned}
  {\boldsymbol{x}^{k+1}} = \argmin_{\boldsymbol{x}}  \gamma\Vert \boldsymbol{y} -\mathrm{M}\boldsymbol{x} \Vert_{2}^2 + \sum_i\Vert(P_i(\boldsymbol{x}) + \frac{\boldsymbol{\lambda}_1^k}{\boldsymbol{\mu}_1^k})\\
  -\Phi\boldsymbol{\alpha}_i^{k+1} \Vert_{2}^2 
  + \frac{\boldsymbol{\mu}_2^k}{2}\Vert (\boldsymbol{x}+\frac{\boldsymbol{\lambda}_2^k}{\boldsymbol{\mu}_2^k}) - \boldsymbol{u}^{k+1} \Vert_{2}^2 \\
\end{aligned}
\end{equation}
A closed-form solution for $\boldsymbol{x}$ exists as follows: \par
\begin{equation} \label{x_closed_form}
\begin{aligned}
  &{\boldsymbol{x}^{k+1}} = (\gamma \mathrm{M}^T\mathrm{M} + \boldsymbol{\mu}_1^k \sum_i {{\rm P}_i}^T {\rm P}_i        +\boldsymbol{\mu}_2^k \mathrm{I})^{-1} \\ 
   &( \gamma \mathrm{M}^T\boldsymbol{y}+ \boldsymbol{\mu}_1^k \sum_i {\rm P}_i\Phi \boldsymbol{\alpha}_i^{k+1} + \boldsymbol{\mu}_2^k \boldsymbol{u}^{k+1} - \sum_i {\rm P}_i \boldsymbol{\lambda}_1^k -  \boldsymbol{\lambda}_2^k \mathrm{I}) \\
\end{aligned}
\end{equation}\par
{In} the expression above, the patch selection operator $P_i(\cdot)$ is rewritten in its matrix form ${\rm P}_i$. To calculate $\boldsymbol{x}^{k+1}$ in practice, {we can leverage the fact that,} for the HSI inpainting task, both $\mathrm{M}^T\mathrm{M}$ and ${{\rm P}_i}^T{\rm P}_i$ are diagonal matrices. {As a results,} the matrix inversion of $(\gamma \mathrm{M}^T\mathrm{M} + \boldsymbol{\mu}_1^k \sum_i {{\rm P}_i}^T{\rm P}_i        +\boldsymbol{\mu}_2^k \mathrm{I})^{-1}$ can be simply implemented as an element-wise division. {This trick significantly reduces the computational cost of the inversion, making the proposed algorithm more scalable for practical use.}\par
{(4)} \textit{{The} Lagrangian and penalty terms are updated as follows}: \\
\begin{equation}
\begin{aligned}
  {\boldsymbol{\lambda}_1^{k+1}} = \boldsymbol{\lambda}_1^{k} + \boldsymbol{\mu}_1^{k}(\boldsymbol{x}^{k+1}-\Phi \boldsymbol{\alpha}^{k+1})\\
   {\boldsymbol{\lambda}_2^{k+1}} = \boldsymbol{\lambda}_2^{k} + \boldsymbol{\mu}_2^{k}(\boldsymbol{x}^{k+1}-\boldsymbol{u}^{k+1})
\end{aligned}
\end{equation}
\begin{equation}
\begin{aligned}
 \boldsymbol{\mu}_1^{k+1} = {\rho}_1 \boldsymbol{\mu}_1^{k}\\
 \boldsymbol{\mu}_2^{k+1} = {\rho}_2 \boldsymbol{\mu}_2^{k}
\end{aligned}
\end{equation} \par

{The} LRS-PnP-DIP algorithm proposed in \cite{li2023self} is summarized in Algorithm \ref{algorithm:LRS-PnP-DIP}.

\begin{algorithm}[H]
    \caption{(LRS-PnP-DIP) algorithm}.
    \label{algorithm:LRS-PnP-DIP}
  \begin{algorithmic}
    \REQUIRE masking matrix: $\mathrm{M}$, noisy and incomplete HSI: $\boldsymbol{y}$, learned dictionary: $\Phi$. denoiser: $\mathcal{D}$. max iteration: $It_{max}$. DIP: $f_{\theta}$
    \OUTPUT inpainted HSI image $\boldsymbol{x}$.
    \STATE \textbf{Initialization} DIP parameters $\theta$, $\boldsymbol{\lambda}_1,\boldsymbol{\lambda}_2,\boldsymbol{\mu}_1,\boldsymbol{\mu}_2, {\rho}_1,{\rho}_2$.
    \WHILE{Not Converged}
      \STATE  for $i=1:It_{max}$ do:
      
               $\boldsymbol{\alpha}^{k+1} = \mathcal{D}(\mathrm{I}-\nabla f)(\boldsymbol{\alpha}^{k})$
      \STATE {update DIP parameters $\theta$, with target: $\boldsymbol{y}$ and intermediate results: $\boldsymbol{x}^k + \frac{\boldsymbol{\lambda}_2^k}{\boldsymbol{\mu}_2^k}$ as input.}
      \STATE update $\boldsymbol{x}$ by \eqref{x_closed_form}.
      \STATE update Lagrangian parameters and penalty terms.
    \ENDWHILE
  \end{algorithmic}
\end{algorithm} 
\noindent {\textit{Remarks.}}
 In contrast to its original form, it is here proposed to replace the conventional PnP denoiser in \eqref{pnp_ista} with an averaged NLM denoiser, {as introduced} in \cite{PnP_Non_Expansive_D}.>>> {Additionally, we modify} the $f_\theta(\boldsymbol{z})$ in \eqref{DIP_replaced} to obtain Lipschitz continuity with Lipschitz constant 1. {The averaged NLM denoiser is a variant of NLM denoiser whose weight is designed to be a doubly symmetric matrix, {ensuring that both the} columns and rows sum to 1. With this doubly symmetric property, the spectral norm of the weight matrix is bounded by 1, meaning the averaged NLM denoiser is by design, non-expansive.} (We refer the reader \mbox{to \cite{PnP_Non_Expansive_D}} for detailed structures of the averaged NLM denoiser).\par
\begin{figure}[H]
  \includegraphics[width=1\textwidth,height=1\textwidth]{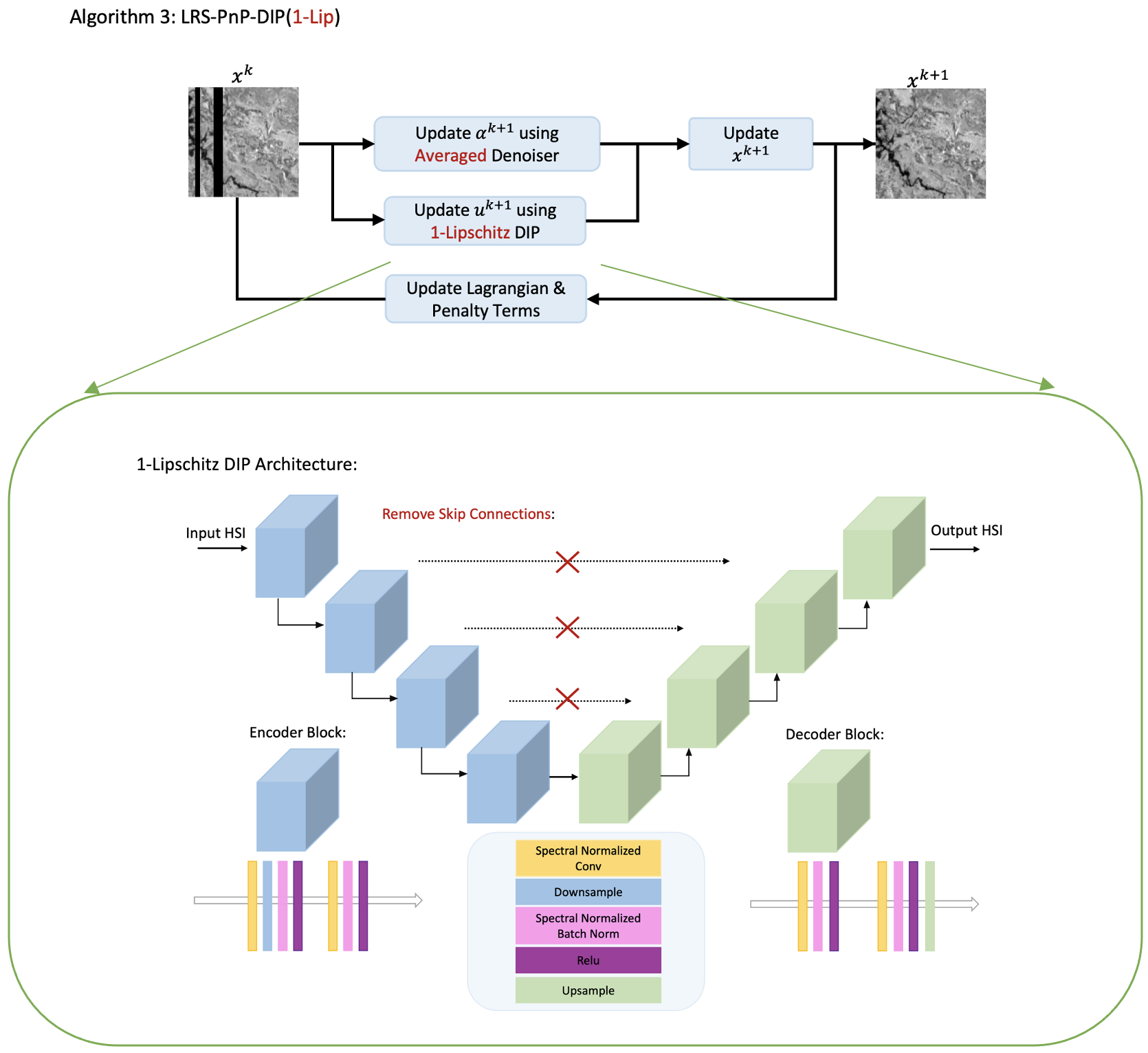}
\caption{{Flow} 
 chart of the proposed LRS-PnP-DIP(1-Lip) algorithm. {At each iteration, $\boldsymbol{\alpha}^{k+1}$ $\boldsymbol{u}^{k+1}$ and $\boldsymbol{x}^{k+1}$ are sequentially updated.} The 1-Lipschitz DIP is implemented by imposing Lipschitz constraints on all layers. {We use red color to highlight the differences between this work and the LRS-PnP-DIP algorithm \cite{li2023self}.} The detailed design of the 1-Lipschitz DIP is placed in Appendix \ref{appendix_1_Lip_DIP}.}
\label{Flow_Chart_LRS_PnP_DIP_1_Lip}  
\end{figure}

{The} enhanced algorithm is denoted as LRS-PnP-DIP(1-Lip), which is shown in \mbox{Algorithm \ref{algorithm:LRS-PnP-DIP_1_lip}}, and its flow chart is provided in Figure \ref{Flow_Chart_LRS_PnP_DIP_1_Lip}.
\vspace{-6pt}

\begin{algorithm}[H]
    \caption{LRS-PnP-DIP(1-Lip) {algorithm.}
}
    \label{algorithm:LRS-PnP-DIP_1_lip}
  \begin{algorithmic}
    \REQUIRE masking matrix: $\mathrm{M}$, noisy and incomplete HSI: $\boldsymbol{y}$, learned dictionary: $\Phi$. \textcolor{red}{averaged NLM Denoiser}: $\mathcal{D}$. max iteration: $It_{max}$. \textcolor{red}{1-Lipschitz DIP}: $f_{\theta}$
    \OUTPUT inpainted HSI image $\boldsymbol{x}$.
    \STATE \textbf{Initialization} DIP parameters $\theta$, $\boldsymbol{\lambda}_1,\boldsymbol{\lambda}_2,\boldsymbol{\mu}_1,\boldsymbol{\mu}_2,{\rho}_1,{\rho}_2$.
    \WHILE{Not Converged}
      \STATE  for $i=1:It_{max}$ do:
      
          $\boldsymbol{\alpha}^{k+1} = \mathcal{D}(\mathrm{I}-\nabla f)(\boldsymbol{\alpha}^{k})$
      \STATE {update DIP parameters $\theta$, with target: $\boldsymbol{y}$ and intermediate results: $\boldsymbol{x}^k + \frac{\boldsymbol{\lambda}_2^k}{\boldsymbol{\mu}_2^k}$ as input.}
      \STATE update $\boldsymbol{x}$ by \eqref{x_closed_form}.
      \STATE update Lagrangian parameters and penalty terms.
    \ENDWHILE
  \end{algorithmic}
\end{algorithm} 

\section{Convergence Analysis}\label{sec3}
In this section, we describe the fixed-point convergence of the proposed LRS-PnP-DIP(1-Lip) algorithm under mild assumptions. Fixed-point convergence refers to the type of convergence wherein the algorithm asymptotically enters a steady state. To establish our main theorem, we need the following results:

\begin{Definition}(Non-expansive operator). An operator, $T:\mathbb{R}^n \rightarrow \mathbb{R}^n$, is said to be non-expansive if, for any $x, y \in \mathbb{R}^n$, the following applies:
\begin{equation}
\begin{aligned}
     \Vert(T(x)-T(y))\Vert^2 &\le  \Vert(x-y)\Vert^2 
\end{aligned}
\end{equation}
\end{Definition}
\begin{Definition}($\theta$-averaged). An operator, $T:\mathbb{R}^n \rightarrow \mathbb{R}^n$, is said to be $\theta$-averaged with some $\theta \in (0,1)$ if there exists a non-expansive operator, $R$, such that we can write $T = (1-\theta) I+\theta R$.
\end{Definition}
\begin{Lemma} \label{Lemma 1}
Let $T:\mathbb{R}^n \rightarrow \mathbb{R}^n$ be $\theta$-averaged for some $\theta \in (0,1)$. Then, for any $x, y \in \mathbb{R}^n$, the following applies:
\begin{equation}
\begin{aligned}
     \Vert(T(x)-T(y))\Vert^2 &\le  \Vert(x-y)\Vert^2 \\ &-\frac{1-\theta}{\theta}\Vert((I-T)(x)-(I-T)(y)\Vert^2
\end{aligned}
\end{equation}
Proof of this lemma can be found in [\cite{nair2021fixed}, Lemma 6.1].
\end{Lemma}
\begin{Definition}(Fixed point). We say that $x^*\in\mathbb{R}^n$ is a fixed point of the operator $T: \mathbb{R}^n \rightarrow \mathbb{R}^n$ if $T(x^*)=x^*$. We denote the set of fixed points as fix($T$). 
\end{Definition} 
\begin{Definition}($\beta$-smoothed). Let $f:\mathbb{R}^n \rightarrow \mathbb{R}^n$ be differentiable; we say that $f$ is $\beta$-smooth if there exists $\beta >0$, such that $\Vert\nabla f(x) - \nabla f(y)\Vert \le \beta \Vert x-y \Vert $ for any $x,y \in \mathbb{R}^n$. 
\end{Definition} 
\begin{Definition}(Strong convexity). A differentiable function, $f$, is said to be strongly convex with modulus $\rho>0$ if $f(x)- \frac{\rho}{2}\Vert x \Vert ^2$ is convex.
\end{Definition}
\begin{Lemma} \label{Lemma 2}
(Property of strong convexity). Let $f$ be strongly convex with modulus $\rho>0$. Then, for any $x, y \in \mathbb{R}^n$, the following applies: \\
\begin{equation} 
\begin{aligned}
     \left \langle \nabla f(x)-\nabla f(y), x-y\right \rangle &\ge  \rho \Vert(x-y)\Vert^2 
\end{aligned}
\end{equation} 
Proof of this Lemma can be found in \cite{ekeland1999convex}. 
\end{Lemma} 
\noindent We made the following assumptions: 
\begin{Assumption} \label{Assump 1}
We assume that (1) the denoiser $\mathcal{D}$ used in the sparse coding step is linear and $\theta$-averaged for some $\theta \in (0, 1)$, (2) the function $f(\boldsymbol{\alpha})=\frac{\boldsymbol{\mu}_1^k}{2} \Vert(\boldsymbol{x}^k + \frac{\boldsymbol{\lambda}_1^k}{\boldsymbol{\mu}_1^k}) - \Phi \boldsymbol{\alpha} \Vert_{2}^2 $ is $\beta$-smoothed, and (3) $ﬁx(\mathcal{D}({\mathrm{I}}-\nabla f)) \neq \emptyset $.
\end{Assumption} 
\noindent \textit{Remarks}. The $\theta$-averaged property is a subset of the non-expansive operator that most of the existing PnP frameworks have worked with \cite{PnP_Non_Expansive_D,PnP_Non_Expansive_Network}. In fact, the non-expansive assumption is found to be easily violated for denoisers such as BM3D and NLM denoisers when used in practice. Nevertheless, the convergence of PnP methods using such denoisers can still be empirically verified. {In the experiments, we adopted the modified NLM denoiser whose spectral norm is bounded by 1. Hence, it satisfies the $\theta$-averaged property by \mbox{design \cite{PnP_Non_Expansive_D}}.}
\begin{Lemma}(Fixed-point convergence of PnP sparse coding). If Assumption \ref{Assump 1} holds, then, for any $\boldsymbol{\alpha}$, and $\rho>\beta/2$, the sequence $\boldsymbol{(\alpha)^k}_{k\ge0}$ generated via step \eqref{pnp_ista} converges to some \mbox{$\boldsymbol{\alpha}^* \in ﬁx(\mathcal{D}(I-\nabla f))$}. \\
Proof of this Lemma can be found in [\cite{nair2021fixed}, Theorem 3.5]. \end{Lemma}
\begin{Assumption} \label{Assump 2}
We assume that the DIP function $f_{\theta}$ is L-Lipschitz-bounded: 
\begin{equation}
\begin{aligned}
     \Vert(f_\theta(x)-f_\theta(y))\Vert^2 \le L \Vert(x-y)\Vert^2 
\end{aligned}
\end{equation}
for any $x$,$y$, and $L\le1$. 
\end{Assumption}
\noindent {\textit{Remarks}.} {In the above assumption, the smallest constant $L$ that makes the inequality hold is called the Lipschitz constant. The Lipschitz constant is expressed as the maximum ratio between the absolute change in the output and the input. It quantifies how much the function output changes with respect to the input perturbations or, roughly speaking, how robust the function $f_\theta$ is. In deep learning, it. Assumption \ref{Assump 2} guarantees that the trained DIP has a Lipschitz constant, $L\le1$, which is a desirable feature so that a network does not vary drastically in response to minuscule changes in its input. In the experiments, this can be achieved by constraining the spectral norm of each layer of the neural network \mbox{during training}. } \par
{We} are now ready to state our main theorem. 
\begin{Theorem} \label{theorem 1} (Convergence of LRS-PnP-DIP(1-Lip) in the Lyapunov Sense). If both Assumption \ref{Assump 1} and Assumption \ref{Assump 2} hold, with penalty $\boldsymbol{\mu}$, and with an L-Lipschitz constrained DIP ($L\le1$), then there exists a non-increasing function, $H^k = 2\Vert \boldsymbol{x}^k-\boldsymbol{x}^*\Vert^2 +\frac{1}{\boldsymbol{\mu}^2}\Vert \boldsymbol{\lambda}_1^k-\boldsymbol{\lambda}_1^*\Vert^2 +\frac{1}{\boldsymbol{\mu}^2}\Vert \boldsymbol{\lambda}_2^k-\boldsymbol{\lambda}_2^*\Vert^2$, such that all trajectories generated via LRS-PnP-DIP(1-Lip) are bounded, and that as $k \rightarrow \infty$, $\Vert \boldsymbol{x}^{k} - \boldsymbol{x}^* \Vert^2 \rightarrow 0$, $\Vert \boldsymbol{\alpha}^{k} -\boldsymbol{\alpha}^* \Vert^2 \rightarrow 0$, and $\Vert \boldsymbol{u}^{k} - \boldsymbol{u}^{*} \Vert^2 \rightarrow 0$; i.e., even if the equilibrium states of $\boldsymbol{x}$, $\boldsymbol{\alpha}$, and $\boldsymbol{u}$ are perturbed, they will finally converge to $\boldsymbol{x}^{*}$, $\boldsymbol{\alpha}^{*}$, and $\boldsymbol{u}^{*}$, respectively. The proposed LRS-PnP-DIP(1-Lip) algorithm is, thus, asymptotically stable. 
\end{Theorem} 
\noindent We placed the detailed proof of Theorem \ref{theorem 1} in Appendix \ref{proof_convergence}.

\section{Experimental Results}\label{sec4}
\subsection{Implementation Details}

We evaluate the proposed inpainting model on two publicly available\linebreak hyperspectral datasets:
\begin{itemize}
    \item The Chikusei airborne hyperspectral dataset \cite{Chikusei} ({the link to the Chikusei dataset can be found at \url{https://naotoyokoya.com/Download.html}, accessed on 4 January 2025}); the test HS image consists of 192 channels, and it was cropped to 36 × 36 pixels in size.
    \item The Indian Pines dataset from AVIRIS sensor \cite{Indian_Pine}; the test HS image consisted of \mbox{200 spectral} bands, and it was cropped to 36 × 36 pixels size.
\end{itemize}\par
{For} each dataset, we trained the dictionary $\Phi$ with a size of 1296 $\times$ 2000, using only noisy and incomplete HSI images {(the size of the test HS image and the dictionary were chosen to match the practical use case where the proposed inpainting algorithms were to be mounted on each nano-satellite to process different small image tiles in parallel
 \cite{denby2020orbital}), this} is a necessary step if we do not have access to pure spectra. If a standard spectral dataset exists, learning can be performed in the form of a pre-training step. However, we used the input HS image for this task to demonstrate that the proposed algorithms operate in a self-supervised setting. {Additionally, we introduced Gaussian noise with noise level $\sigma = 0.12$ to the cropped HS images.} {Since the main focus of this paper is image inpainting, instead of denoising, $\sigma$ was kept fixed across all experiments}. We applied mask $\mathrm{M}$ to all the spectral bands in the given region, which represents the most challenging case. For the choice of the PnP denoiser, we used both the BM3D and modified NLM denoisers \cite{PnP_Non_Expansive_D}; the latter is an averaged denoiser, which enjoys a non-expansive property that is crucial to convergence analysis.\par
{Our} implementation of the deep image prior model follows the same structure as in the deep hyperspectral prior \cite{sidorov2019deep}. {During the training of} 1-Lipschitz DIP, we applied Lipschitz regularization to all layers, as in \cite{henry_gouk_2021regularisation}. {In each iteration, we fed the DIP with the latent image from the previous iteration, rather than the random signal, as suggested by \cite{sidorov2019deep}.} We did not perturb the input to support our convergence analysis, {which is different from the canonical DIP. To eliminate the need for manually selecting the total number of iterations for the DIP model,} we adopted the early stopping criterion proposed in \cite{wang2021early} with a fixed window size of 20 and a patience number of 100, which detects the near-peak MPSNR point using windowed moving variance (WMV). {Note that, for the training of conventional DIP, the early-stopping strategy is often employed to prevent it from fitting to random noise after certain iterations. However, the proposed LRS-PnP-DIP(1-Lip) algorithm mitigates the over-fitting issue due to the imposed constraint. The early-stopping criterion is applied primarily to optimize the algorithm's runtime so that one does not have to wait for the algorithm to reach the maximum number of iterations. In our formulation,} the low-rank and sparsity constraints in the equation were weighted with parameters $w_{lr}$ and $w_{s}$, respectively, while the data fidelity term was weighted with $\gamma$. Initially, we set $w_{s}/w_{lr}$ to 1 and $\gamma$ to 0.5. Notably, the choice of $\gamma$ is highly dependent on the noise level of the observed image. If the noise level is low, the recovered image, $X$, should be similar to the noisy observation, $Y$, parameter $\gamma$ should be large, and vice versa. The Adam optimizer was used, and the learning rate was set to 0.1. {Finally, we used three widely used metrics, namely the mean signal-to-noise ratio (MPSNR), mean structural similarity (MSSIM), and mean spectral angle mapper (MSAM), to evaluate the performance of the model in all experiments.}

\subsection{Low-Rank vs. Sparsity}\label{seiton Low-Rank v.s. Sparsity}
\begin{figure}[H]

\includegraphics[width=0.82\textwidth,height=0.6\textwidth,]{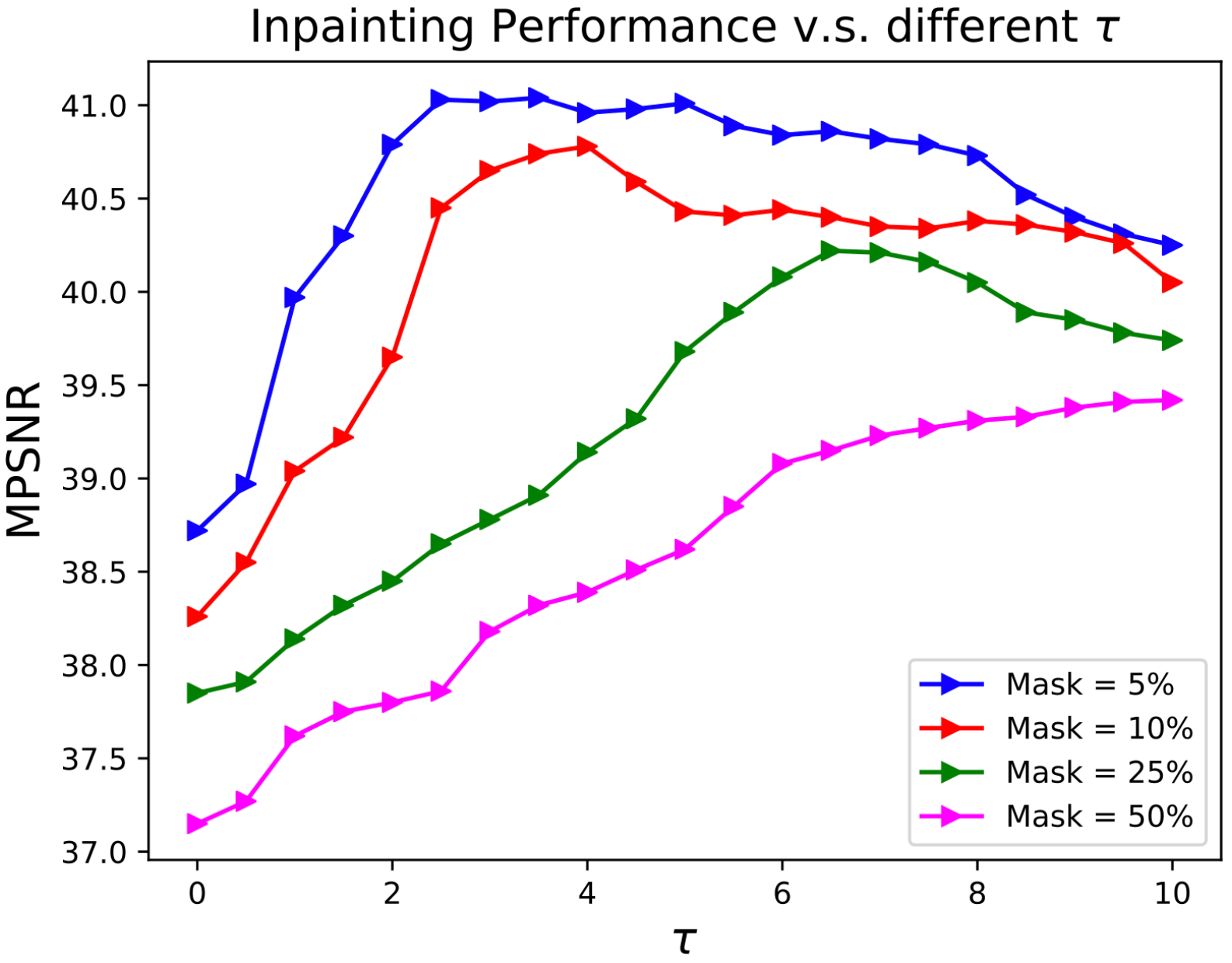} 
\caption{Comparison of MPSNR value of LRS-PnP among different $\tau$ ($\tau = w_{s}/w_{lr}$) under \mbox{different masks}.}
 \label{fig: sparsity vs low-rank}
\end{figure}

In this section, we examine the effectiveness of both sparsity and low-rank constraints in our proposed LRS-PnP algorithm. {Figure \ref{fig: sparsity vs low-rank} shows the inpainting performance of LRS-PnP over different $\tau$ for different masks (in percentages), where $\tau$ is defined as the ratio of the weight of the sparsity constraint and the low-rank constraint.} For example, $\tau=0$ means that the sparsity constraint is disabled, and mask = 50\% means that 50\% of the pixels are missing. We can conclude from the results that (1) the LRS-PnP algorithm works better with both low-rank and sparsity constraints, as using either an overwhelming sparsity constraint or a low-rank constraint alone tends to reduce inpainting performance, and (2) when the percentage of missing pixels increases, more weight should be placed on the sparsity constraint in order to achieve high-quality reconstruction. On the one hand, approximation errors may be introduced in the sparse coding step; i.e., step \eqref{alpha} in \mbox{Algorithms \ref{algorithm:LRS-PnP-DIP} and \ref{algorithm:LRS-PnP-DIP_1_lip}}. These errors can potentially be suppressed by the constraint of low rank. On the other hand, the sparsity constraint exploits the spectral correlations; i.e., the performance will be enhanced if spatial information is considered. 
\vspace{-6pt}

\subsection{Low Rankness Due to DIP} 
\begin{figure}[H]
 \includegraphics[width=0.82\textwidth,height=0.63\textwidth]{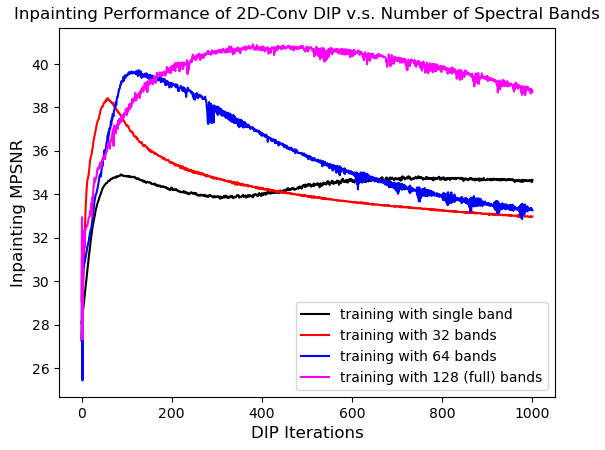}   
 \caption{Learning capability of DIP vs. number of input and output channels. Training is conducted with a single-band HS image, meaning that the input HSIs are processed; there is no correlation in the spectral domain. {There is a significant performance gain when there are more input bands, indicating that the DIP with 2D convolution has the ability to exploit the correlation between channels.}}
 \label{fig: learn_low_rank}
\end{figure}

In the original LRS-PnP-DIP Algorithm \cite{li2023self}, we propose solving the low-rank minimization problem using DIP instead of the traditional SVT projection. A natural question arises: does the DIP (e.g., the architecture proposed in Figure \ref{Flow_Chart_LRS_PnP_DIP_1_Lip}) effectively capture the redundancies in the spectral domain? This is empirically justified in Figure \ref{fig: learn_low_rank}, where we trained DIP on the corrupted HS image with varying numbers of spectral bands as input and output. {The results show that the inpainting performance of DIP improved substantially when more spectral bands were utilized during training; the peak MPSNR was obtained when the neural network saw the entire 128 bands of the input HS images, rather than each single band individually. } This finding is counterintuitive since DIP with 2D convolution is generally expected to handle only the spatial correlation of input HSIs. Similar observations have been reported in \cite{ulyanov2018deep,low_rank_DIP}, where the authors confirmed that DIP with 2D convolution implicitly explored the channel correlations, possibly due to the upsampling layers in the decoder part that manipulates the HSI channels in the spectral domain.\\
\vspace{-18pt}
 \par
\begin{figure}[H]
 \includegraphics[width=0.85\textwidth,height=0.63\textwidth]{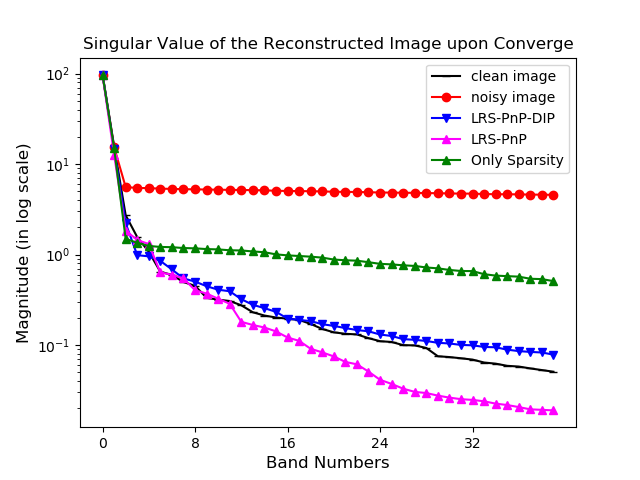}   
 \caption{The amplitude of the singular value of the reconstructed image upon converge. The important singular values are captured and preserved via the 2D-convolution DIP, which is even more accurate than the traditional SVT projection.}
 \label{fig: singular_value}
\end{figure} 
{In} Figure \ref{fig: singular_value}, {we sort the singular values of the inpainted image of LRS-PnP (the pink line), LRS-PnP-DIP \cite{li2023self} (the blue line), and PnP sparse coding without a low-rank constraint (green line), in descending order. Among these algorithms, LRS-PnP performs singular value projection to promote low rankness, while it is replaced with DIP in LRS-PnP-DIP}. It is observed that using a well-trained DIP alone was sufficient to capture these low-rank details. Thus, we suggest replacing the conventional SVT process with DIP. Although there is no theoretical evidence on how DIP mimics the SVT, {we show through experiments that the DIP can well preserve some small yet important singular values that would otherwise be discarded via SVT projection. Hence, it has the ability to better exploit the low-rank subspace in a data-driven fashion, possibly owing to its highly non-linear network structures and inductive bias.

\subsection{Convergence} 
The use of the PnP denoiser and DIP in the context of HSIs has been recently investigated in \cite{lai2022deep, 2021_DIP_In_Loop}. However, theoretical evidence of the stability of the algorithm is still lacking. To bridge this gap, we verify the convergence of LRS-PnP-DIP(1-Lip). In Figure \ref{fig: emperical convergence}, the top left, top right, and bottom left figures represent the successive differences of each state on a log scale, and the bottom right figure represents the inpainting performance. For LRS-PnP-DIP, we used the BM3D denoiser and DIP without constraints. For LRS-PnP-DIP(1-Lip), we used the modified NLM denoiser and the 1-Lipschitz constraint DIP. The latter automatically satisfies Assumptions \ref{Assump 1} and \ref{Assump 2}; thus, the fixed-point convergence can be guaranteed. Interestingly, we noticed that LRS-PnP-DIP with BM3D and conventional DIP still works pretty well in practice even though there are fluctuations in state $\boldsymbol{x}$, as can be seen in the top right and bottom left plots in Figure \ref{fig: emperical convergence}. We deduced that such instability mainly comes from the process of solving DIP sub-problems, rather than the denoising sub-problems, as the primal variable $\boldsymbol{\lambda}_1$ smoothly converges with BM3D denoisers as well. \par

\begin{figure}[H]
\includegraphics[width=0.95\textwidth,height=0.92\textwidth]{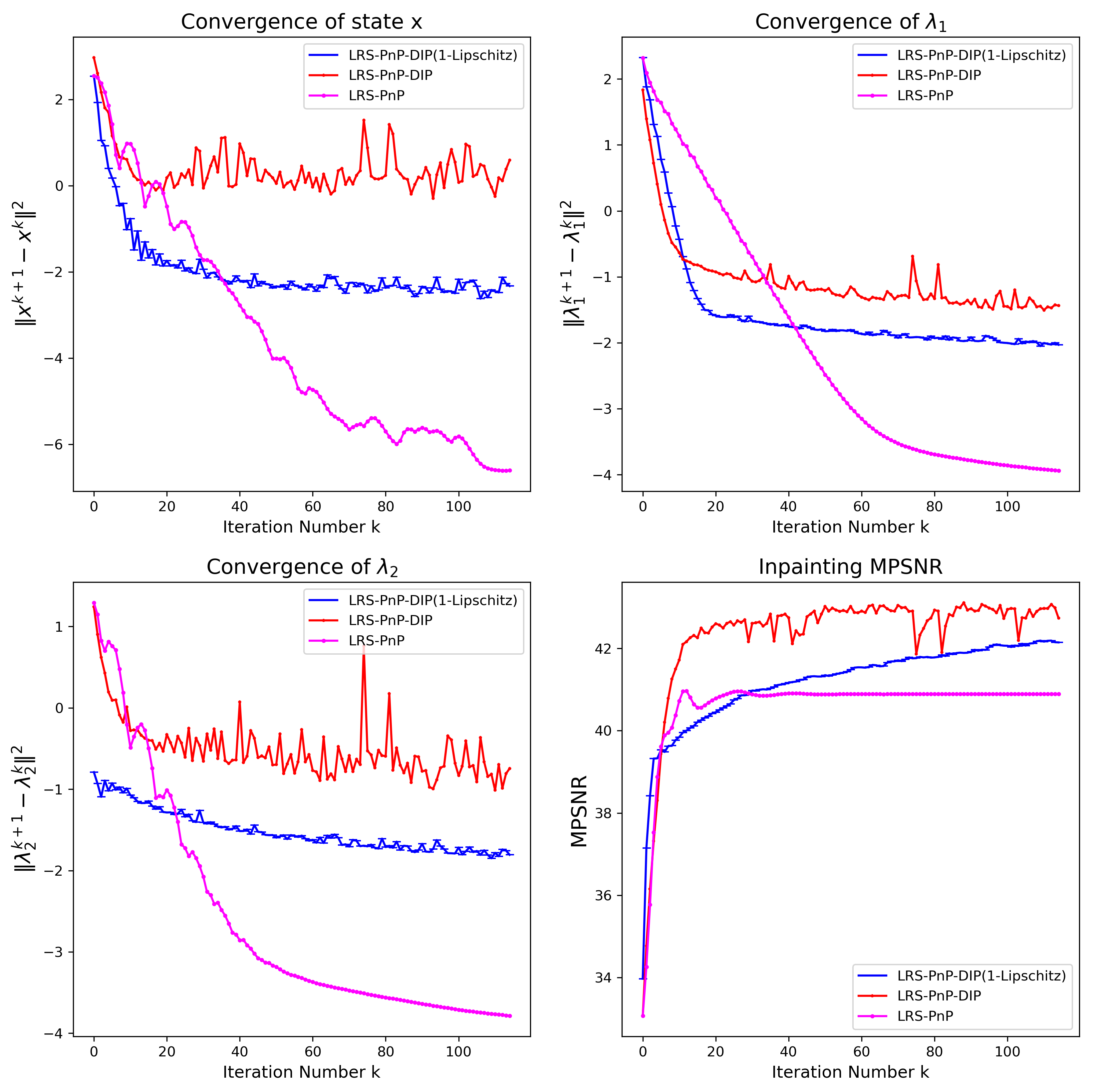} 
 \caption{{Empirical} 
 converge of LRS-PnP-DIP(1-Lip) with modified NLM denoiser and non-expansive/1-Lipschitz DIP. Top left, top right, and bottom left: successive difference of $\boldsymbol{x}$, $\boldsymbol{\lambda}_1$, and $\boldsymbol{\lambda}_2$ in the log scale, respectively. Bottom right: the inpainting MPSNR vs. the number of iterations.}
\label{fig: emperical convergence}
\end{figure}
\vspace{-2pt}
{From} the bottom right plot, we observed that the best results are obtained with LRS-PnP-DIP in terms of reconstruction quality, and the LRS-PnP-DIP(1-Lip) is only slightly lower than its unconstrained counterpart. Our observations on the reduced performance of Lipschitz DIP are in agreement with existing works \cite{miyato2018spectral,henry_gouk_2021regularisation}, where imposing the Lipschitz constraint during training does hurt the capacity and expressivity of the neural network.

\subsection{Comparison with State of the Art}\label{section_compare_others}
The proposed algorithms were compared with learning-based algorithms: DHP \cite{sidorov2019deep}, GLON \cite{zhao2021tensor}, R-DLRHyIn \cite{niresi2023robust}, DeepRED \cite{2019_Deep_Red}, PnP-DIP \cite{pnp_dip},{DeepHyIn \cite{DeepHyIn}, and \mbox{DDS2M \cite{DDS2M}}.} {To ensure the fairness of comparison, we employed the same U-net as the backbone for DHP, DeepRED, R-DLRHyIn, PnP-DIP, and LRS-PnP-DIP(1-Lip) and the attention-based U-net with the same complexity for DeepHyIn and DDS2M.}
For GLON, we replaced its original FFDNet with the one trained on the hyperspectral dataset \cite{maffei2019single} and kept the others intact. {({The reason for making this adaption is that the neural network proposed in GlON \cite{zhao2021tensor} was trained on RGB/grayscale images. It was found to be beneficial to use an FFDNet, \mbox{e.g., in \cite{maffei2019single}}, trained on the HSI dataset}.)}
{For DeepHyIn, we used the end members extracted from the clean image patch. For DDS2M, we fine-tuned the diffusion-related parameters, as suggested in \cite{DDS2M}. The statistical results of each test sample are averaged over 20 experiments to account for the effect of random seeds during training.} \par
\begin{figure}[H]
\hspace{-1.3cm}\includegraphics[width=0.86\textwidth,height=0.62\textwidth,]{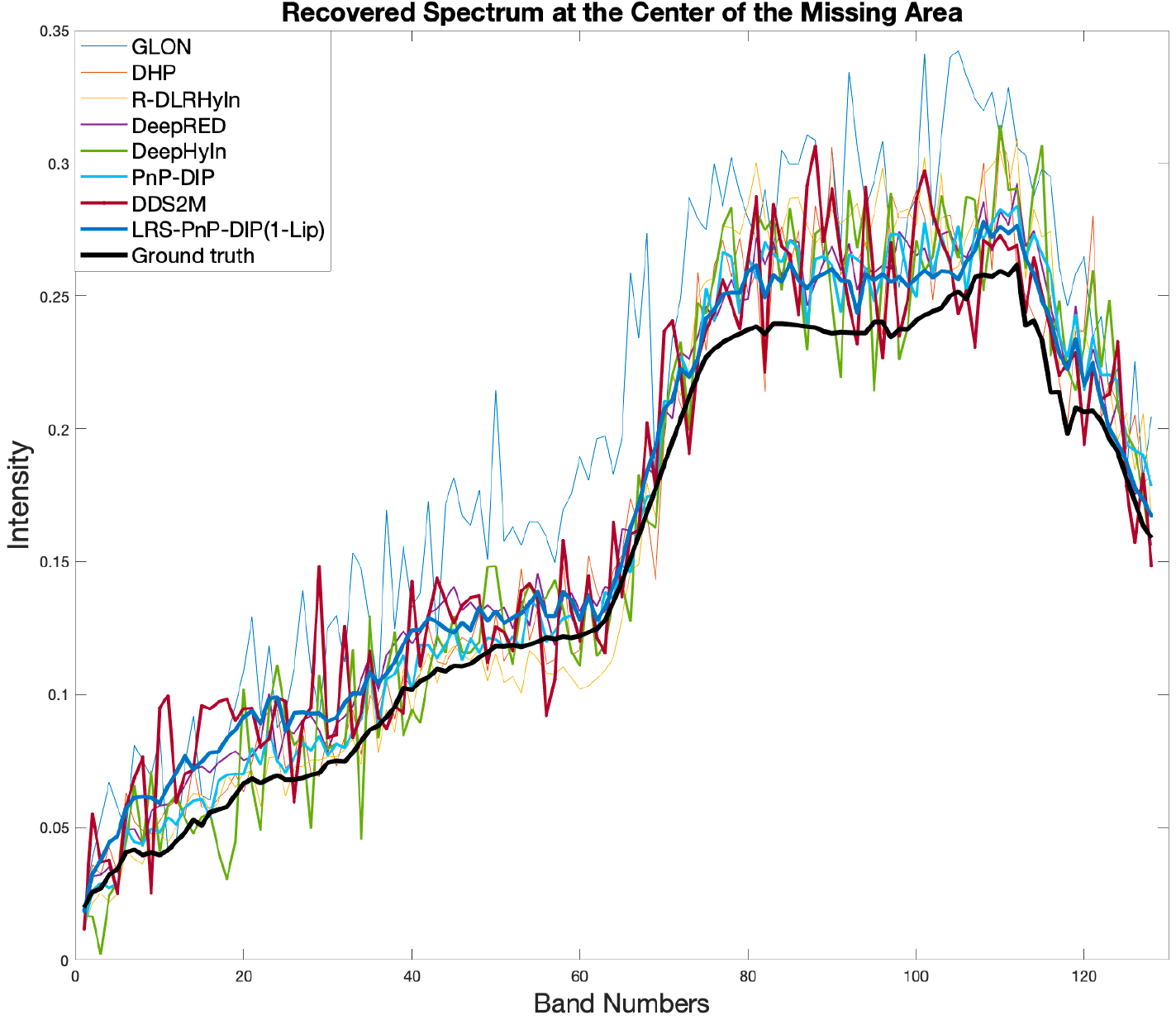} 
\caption{{Different algorithms and their recovered spectrum of the center pixel with the assumption that the whole spectrum bands are missing. MPSNR increases from the thin line to the thick line.}}
\label{recovered spectrum}
\end{figure} \par
{In} Figure \ref{recovered spectrum}, we plot the recovered spectrum of the center of the missing area of LRS-PnP-DIP(1-Lip) for each method. It can be seen that LRS-PnP-DIP(1-Lip) produces a more consistent and realistic spectrum in the missing region compared to methods such as GLON, DHP, DeepRED, R-DLRHyIn, {DDS2M, and DeepHyIn.} {We present the visual inpainting results for each method on the Chikusei dataset and the Indian Pines dataset, as depicted in  Figures \ref{reconstruction results} and \ref{reconstruction results_Indian_Pine}, respectively. The quantitative statistical results for each dataset, evaluated under various mask shapes, are summarized in Table \ref{compare DL methods} for the Chikusei dataset and Table \ref{compare DL methods_Indian_Pines} for the Indian Pines dataset. }

\begin{figure}[H]
\includegraphics[width=1\textwidth,height=1.2\textwidth]{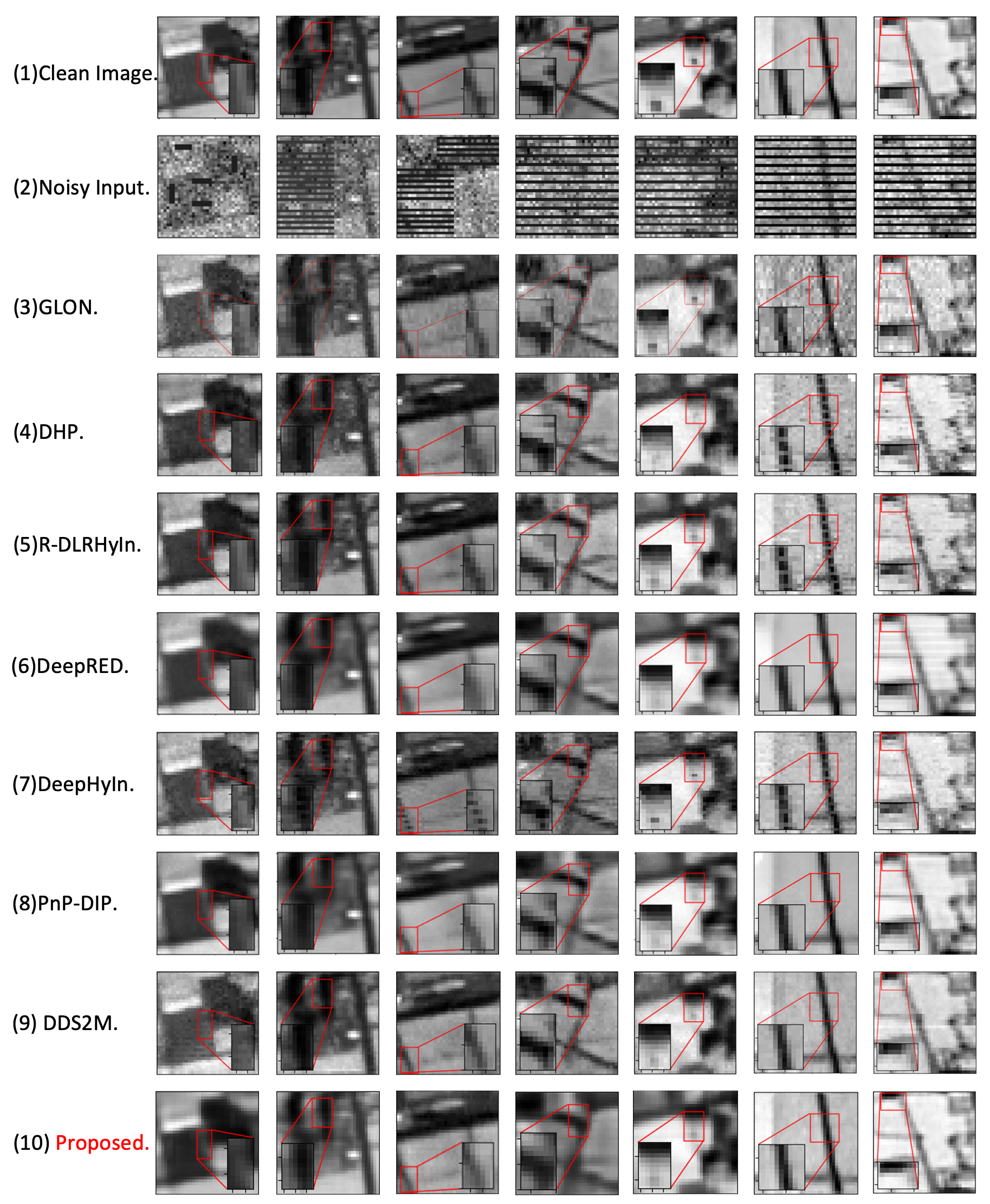}
  \caption{{{Comparison} 
 between the proposed algorithm and other learning-based inpainting algorithms on the Chikusei dataset.} From top to bottom: (\textbf{1}) clean image, (\textbf{2}) input image, (\textbf{3}) GLON, \mbox{(\textbf{4}) DHP}, (\textbf{5}) R-DLRHyIn, (\textbf{6}) DeepRED, (\textbf{7}) {DeepHyIn}, (\textbf{8}) {PnP-DIP}, (\textbf{9}) {DDS2M}, and (\textbf{10}) {LRS-PnP-DIP(1-Lip)}. All images are visualized at band 80.}
  \label{reconstruction results}  
\end{figure}

\begin{table}[H]
\tablesize{\tiny}
\caption{{Comparison} 
 between the proposed algorithm and other learning-based inpainting algorithms on the Chikusei dataset: the mean and variance over 20 samples are shown here. During the experiments, four different mask shapes were used, which are depicted in the first four columns of Figure \ref{reconstruction results}. The best results are underlined for each type of mask.\label{compare DL methods} {For the metrics MPSNR and MSSIM, higher values indicate better performance. Conversely, for the MSAM metric, the lower the better.}}
	\begin{adjustwidth}{-\extralength}{0cm}
		\begin{tabularx}{\fulllength}{crrrrrrrrrr}
			\toprule
\textbf{Mask Type} & \textbf{Methods} & \textbf{Input} & \textbf{GLON} \cite{zhao2021tensor} & \textbf{DHP} \cite{sidorov2019deep} & \textbf{R-DLRHyIn} \cite{niresi2023robust} & \textbf{DeepRED} \cite{2019_Deep_Red}  & \textbf{DeepHyIn} \cite{DeepHyIn}  &\textbf{PnP-DIP} \cite{pnp_dip} &\textbf{DDS2M} \cite{DDS2M} &\textbf{LRS-PnP-DIP(1-Lip)}\\
			\midrule
Mask Type 1 & MPSNR $\uparrow$ & 33.0740 & $ 41.0324$ & 41.3956($\pm$0.52) & 41.5714($\pm$0.31) & 41.5765($\pm$0.28) &41.6902($\pm$0.57) & 41.7595($\pm$0.25)  & ${41.7728(\pm 0.35)}$ & $\uuline{41.8023(\pm0.16)}$ \\ 
\midrule 
\hspace{2cm} & MSSIM $\uparrow$ & 0.2441 & $ 0.9078 $ & 0.9102($\pm$0.03)  & 0.9135($\pm$0.01) & 0.9121($\pm$0.02) & 0.9180($\pm$0.02) & 0.9250($\pm$0.02) & ${0.9268(\pm 0.02)}$ & $\uuline{0.9275(\pm 0.01)}$\\ 
\midrule 
\hspace{2cm} & MSAM $\downarrow$ & 0.7341 & 0.1342 & 0.1296($\pm$0.01)  &0.1250($\pm$0.01) & 0.1237($\pm$0.01) & 0.1221($\pm$0.01) & ${0.1203(\pm0.01)}$ & $\uuline{0.1183(\pm 0.01)}$ & ${ 0.1185(\pm 0.01)}$\\
\midrule

Mask Type 2 & MPSNR $\uparrow$ & 31.7151 & 38.4103 & 39.7510($\pm$0.65) & 40.1586($\pm$0.84) & 40.4119($\pm$0.33) &40.6260($\pm$0.40) &40.6661($\pm$0.52) &40.7904($\pm$0.74) & $\uuline{41.4910(\pm 0.23)}$ \\ 
\midrule 
\hspace{2cm} & MSSIM $\uparrow$ & 0.2338 & 0.8768 & 0.8961($\pm$0.01)  & 0.9041($\pm$0.01) & 0.9049($\pm$0.01) &0.9100($\pm$0.01) & 0.9121($\pm$0.01) &0.9130($\pm$0.01) & $\uuline{0.9189(\pm 0.01)}$\\ 
\midrule
\hspace{2cm} & MSAM $\downarrow$ & 0.8296 & 0.1518 & 0.1436($\pm$0.02)  &0.1373($\pm$0.01) & 0.1320($\pm$0.01) & 0.1315($\pm$0.01) & 0.1298($\pm$0.01) & ${0.1264(\pm 0.01)}$ & $\uuline{ 0.1203(\pm 0.01)}$\\
\midrule

Mask Type 3 & MPSNR $\uparrow$ & 30.3258 & 35.6927 & 38.1440($\pm$0.90) & 38.9580($\pm$0.79) & 39.6952($\pm$0.99) &39.8608($\pm$0.68) &39.8150($\pm$0.40) &40.0779($\pm$0.82) & $\uuline{40.8400(\pm 0.49)}$ \\ 
\midrule 
\hspace{2cm} & MSSIM $\uparrow$ & 0.2018 & 0.8558 &  0.8805($\pm$0.01)  &  0.8871($\pm$0.01) &0.8902($\pm$0.01) &08950($\pm$0.01) & 0.8958($\pm$0.01) &0.9003($\pm$0.01) & $\uuline{0.9088(\pm 0.01)}$\\ 
\midrule
\hspace{2cm} & MSAM $\downarrow$ & 0.8441 & 0.1670 & 0.1563($\pm$0.01)  &0.1507($\pm$0.01) & 0.1495($\pm$0.01) & 0.1480($\pm$0.01) & 0.1476($\pm$0.01) & ${0.1455(\pm 0.01)}$ & $\uuline{ 0.1392(\pm 0.01)}$\\
\midrule

Mask Type 4 & MPSNR $\uparrow$ & 27.9802 & 34.5011 & 36.6018($\pm$1.15) & 37.6504($\pm$0.89) & 37.9080($\pm$0.59) &38.1961($\pm$0.77) &38.3442($\pm$0.73) &38.7529($\pm$0.88) & $\uuline{39.4068(\pm 0.54)}$ \\ 
\midrule 
\hspace{2cm} & MSSIM $\uparrow$ & 0.1700 & 0.8267 & 0.8693($\pm$0.01)  & 0.8740($\pm$0.01) &0.8772($\pm$0.01) & 0.8769($\pm$0.01) & 0.8820($\pm$0.01) &0.8849($\pm$0.01) & $\uuline{0.8927(\pm 0.01)}$\\ 
\midrule
\hspace{2cm} & MSAM $\downarrow$ & 0.8847 & 0.1803 & 0.1714($\pm$0.03)  &0.1636($\pm$0.02) & 0.1602($\pm$0.01) & 0.1580($\pm$0.01) & 0.1557($\pm$0.01) & ${0.1532(\pm 0.01)}$ & $\uuline{0.1449(\pm 0.01)}$\\
			\bottomrule
		\end{tabularx}
	\end{adjustwidth}
\end{table}

\begin{table}[H]
\caption{{Comparison} 
 between the proposed algorithm and other learning-based inpainting algorithms on the Indian Pines dataset: The mean and variance over 20 samples are shown here. During the experiments, four different mask shapes were used, which are depicted in the first four columns of Figure \ref{reconstruction results}. The best results are underlined for each type of mask.\label{compare DL methods_Indian_Pines}}
	\begin{adjustwidth}{-\extralength}{0cm}
		\begin{tabularx}{\fulllength}{crrrrrrrrrr}
			\toprule
\textbf{Mask Type} & \textbf{Methods} & \textbf{Input} & \textbf{GLON} \cite{zhao2021tensor} & \textbf{DHP} \cite{sidorov2019deep} & \textbf{R-DLRHyIn} \cite{niresi2023robust} & \textbf{DeepRED} \cite{2019_Deep_Red}  & \textbf{DeepHyIn} \cite{DeepHyIn}  &\textbf{PnP-DIP} \cite{pnp_dip} & \textbf{DDS2M} \cite{DDS2M} &\textbf{LRS-PnP-DIP(1-Lip)}\\
			\midrule
Mask Type 1 & MPSNR $\uparrow$ & 31.6903 & $ 39.7676$ & 40.1917($\pm$0.33) & 40.5490($\pm$0.41) & 40.7982($\pm$0.32) &40.9309($\pm$0.60) & 40.8897($\pm$0.34)  & ${41.1084(\pm 0.29)}$ & $\uuline{41.3423(\pm 0.24)}$ \\ 
\midrule 
\hspace{2cm} & MSSIM $\uparrow$ & 0.2268 & $ 0.8884 $ & 0.8917($\pm$0.02)  & 0.8930($\pm$0.02) & 0.8932($\pm$0.01) & 0.9028($\pm$0.01) & 0.9050($\pm$0.01) & ${0.9092(\pm 0.01)}$ & $\uuline{0.9139(\pm 0.01)}$\\ 
\midrule 
\hspace{2cm} & MSAM $\downarrow$ & 0.7723 & 0.1594 & 0.1580($\pm$0.01)  &0.1369($\pm$0.01) & 0.1368($\pm$0.01) & 0.1320($\pm$0.01) & ${0.1255(\pm0.01)}$ & ${0.1240(\pm 0.01)}$ & $\uuline{ 0.1256(\pm 0.01)}$\\
\midrule

Mask Type 2 & MPSNR $\uparrow$ & 29.6815 & 36.9092 & 38.1331($\pm$0.22) & 38.7072($\pm$0.25) & 38.5630($\pm$0.19) &38.7976($\pm$0.40) &39.0021($\pm$0.32) &39.9861($\pm$0.40) & $\uuline{40.3010(\pm 0.23)}$ \\ 
\midrule 
\hspace{2cm}& MSSIM $\uparrow$ & 0.2098 & $ 0.8457 $ & 0.8849($\pm$0.01)  & 0.8902($\pm$0.01) & 0.8864($\pm$0.01) & 0.8919($\pm$0.01) & 0.8972($\pm$0.01) & ${0.9011(\pm 0.01)}$ & $\uuline{0.9048(\pm 0.01)}$\\ 
\midrule
\hspace{2cm} & MSAM $\downarrow$ & 0.8723 & 0.1514 & 0.1509($\pm$0.01)  &0.1469($\pm$0.01) & 0.1450($\pm$0.01) & 0.1444($\pm$0.01) & ${0.1420(\pm0.01)}$ & ${0.1380(\pm 0.01)}$ & $\uuline{ 0.1310(\pm 0.01)}$\\
\midrule

Mask Type 3 & MPSNR $\uparrow$ & 28.6195 & 36.7447 & 37.0714($\pm$0.28) & 37.4790($\pm$0.20) & 37.3976($\pm$0.39) &37.4555($\pm$0.39) &37.7821($\pm$0.22) &37.8070($\pm$0.52) & $\uuline{38.0599(\pm 0.33)}$ \\ 
\midrule 
\hspace{2cm}& MSSIM $\uparrow$ & 0.1864 & $ 0.8395 $ & 0.8787($\pm$0.01)  & 0.8870($\pm$0.01) & 0.8869($\pm$0.01) & 0.8893($\pm$0.01) & 0.8921($\pm$0.01) & ${0.8945(\pm 0.01)}$ & $\uuline{0.8982(\pm 0.01)}$\\ 
\midrule
\hspace{2cm} & MSAM $\downarrow$ & 0.8823 & 0.1584 & 0.1579($\pm$0.01)  &0.1539($\pm$0.01) & 0.1521($\pm$0.01) & 0.1490($\pm$0.01) & ${0.1488(\pm0.01)}$ & ${0.1470(\pm 0.01)}$ & $\uuline{ 0.1458(\pm 0.01)}$\\
\midrule

Mask Type 4  & MPSNR $\uparrow$ & 24.6998 & 34.5630 & 35.8009($\pm$0.30) & 35.8292($\pm$0.22) & 35.8145($\pm$0.32) &35.8905($\pm$0.23) &36.1022($\pm$0.18) &36.4938($\pm$0.34) & $\uuline{36.5104(\pm 0.23)}$ \\ 
\midrule 
\hspace{2cm}& MSSIM $\uparrow$ & 0.1672 & $ 0.8342 $ & 0.8592($\pm$0.01)  & 0.8651($\pm$0.01) & 0.8678($\pm$0.01) & 0.8698($\pm$0.01) & 0.8733($\pm$0.01) & ${0.8764(\pm 0.01)}$ & $\uuline{0.8779(\pm 0.01)}$\\ 
\midrule
\hspace{2cm} & MSAM $\downarrow$ & 0.8977 & 0.1644 & 0.1610($\pm$0.02)  &0.1611($\pm$0.01) & 0.1602($\pm$0.01) & 0.1588($\pm$0.01) & ${0.1580(\pm0.01)}$ & $\uuline{0.1552(\pm 0.01)}$ & ${ 0.1554(\pm 0.01)}$\\
			\bottomrule
		\end{tabularx}
	\end{adjustwidth}
\end{table}

\begin{figure}[H]
\includegraphics[width=0.97\textwidth,height=1.35\textwidth]{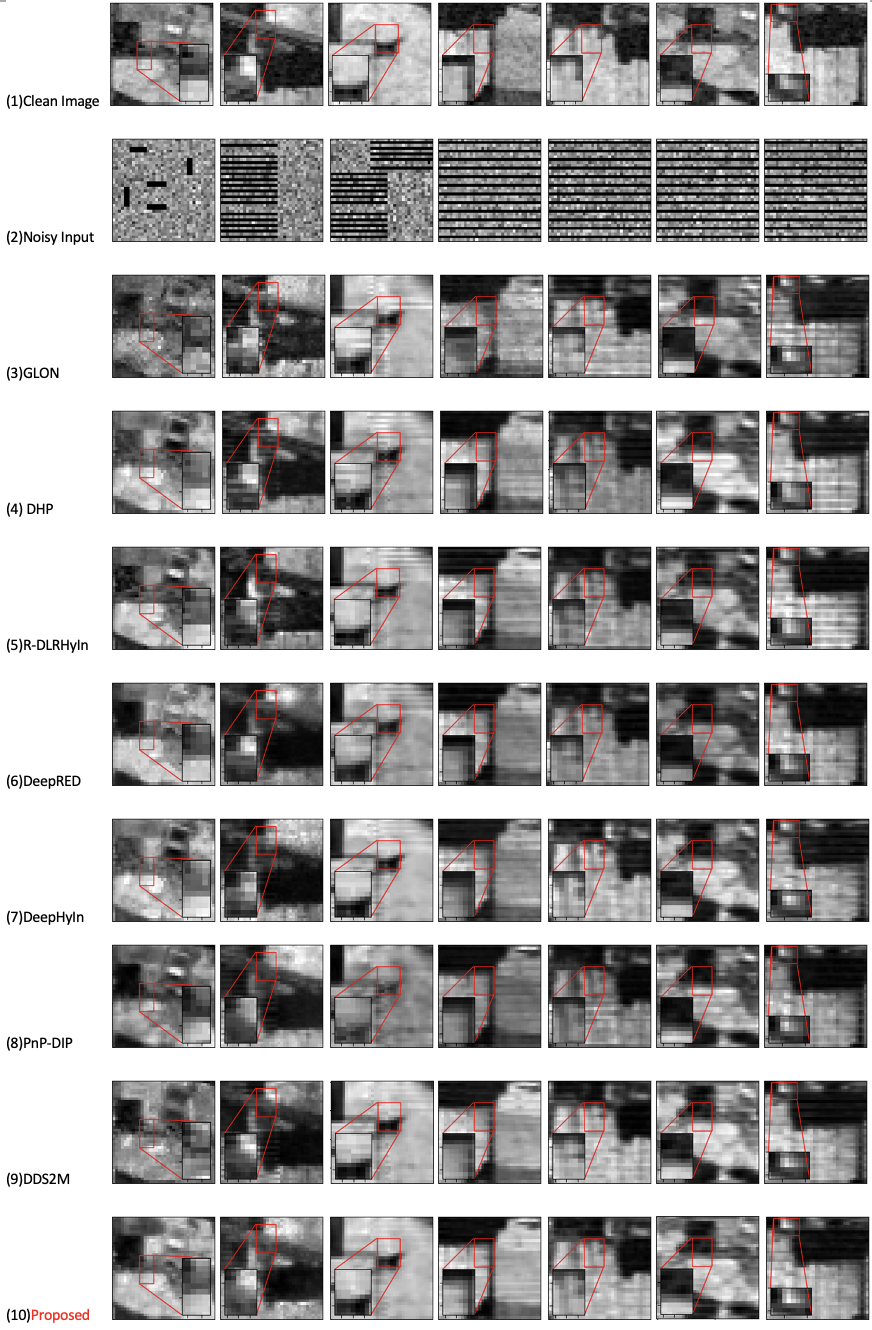}
  \caption{{{Comparison} 
 between the proposed algorithm and other learning-based inpainting algorithms on the Indian Pines dataset. From top to bottom: (\textbf{1}) clean image, (\textbf{2}) input image, (\textbf{3}) GLON, (\textbf{4}) DHP, (\textbf{5}) R-DLRHyIn, (\textbf{6}) DeepRED, (\textbf{7}) {DeepHyIn}, (\textbf{8}) {PnP-DIP}, (\textbf{9}) {DDS2M}, and (\textbf{10}) {LRS-PnP-DIP(1-Lip)}. All images are visualized at band 150.}}
  \label{reconstruction results_Indian_Pine}  
\end{figure} \par

{It} is shown that LRS-PnP-DIP(1-Lip) can effectively suppress noise compared to methods such as R-DLRHyIn, DeepHyIn, and DDS2M. Moreover, it alleviates the over-smoothing phenomenon of the DeepRED and PnP-DIP algorithms by preserving more details in the non-missing region; this is clear for the last two test samples in Figure \ref{reconstruction results} {and for the first four test samples in Figure \ref{reconstruction results_Indian_Pine}. From Tables \ref{compare DL methods} and \ref{compare DL methods_Indian_Pines}, it can be seen that LRS-PnP-DIP(1-Lip) almost always demonstrates the highest inpainting performance on both the Chikusei and Indian Pines datasets compared to other methods. The second best is established via DDS2M, which is the closest method to ours in terms of inpainting quality. This is possibly due to its ability to leverage inductive bias from hierarchical diffusion structures. However, as is evident in Figures \ref{reconstruction results} and \ref{reconstruction results_Indian_Pine}, DDS2M and other DIP-based extensions occasionally exhibit limitations, such as producing noisy backgrounds or failing to effectively inpaint missing strips. It is important to note that these methods utilize conventional DIP as backbones; hence, they are susceptible to the instability issue of DIP where the algorithm may eventually fit into pure noise without proper constraints. In contrast, LRS-PnP-DIP(1-Lip) is superior, as it not only harvests the strong learning capability of DHP but also enjoys a stability guarantee.} For the HSI inpainting task, we anticipate that some information, from the background/non-missing pixels to be carried to the deeper layers, there is a performance loss when removing the skip connections between the encoder and the decoder. Nevertheless, LRS-PnP-DIP(1-Lip) mutes these layers to trade performance for convergence, and it is empirically found that, even if LRS-PnP-DIP(1-Lip) is highly restricted due to layer-wise spectral normalization, it can preserve more textures and produce less noisy results than DDS2M. {This observation leads to the conclusion that the proposed LRS-PnP-DIP(1-Lip) algorithm would be preferable for solving practical inpainting tasks compared to existing solutions. It has the ability to yield faithful reconstructions while ensuring stability and robustness.}

\section{Conclusions}\label{sec5}
The LRS-PnP algorithm, along with its variant LRS-PnP-DIP proposed in the previous work \cite{li2023self}, are innovative hyperspectral inpainting methods designed to address the challenge of missing pixels in noisy and incomplete HS images. {These techniques can better handle the challenging situations in practical HSI acquisition systems where the entire spectral band may be absent. Through experiments, we show that both sparsity and the low-rank priors can be effectively integrated as constraints during the training of DIP; the resulting LRS-PnP-DIP algorithm leverages the strong learning capability of the powerful generative DIP model and is able to exploit spectral and spatial redundancies inherent to HSIs. Because the proposed methods do not require any training data, they are particularly attractive for practical hyperspectral inpainting tasks, where only the observed corrupted image is available for reference. }
However, the LRS-PnP-DIP algorithm suffers from instability issues and may occasionally diverge due to the DIP. In this paper, this long-lasting issue was effectively addressed through the proposed LRS-PnP-DIP(1-Lip) algorithm. Experimental results on real datasets demonstrate that the LRS-PnP-DIP(1-Lip) algorithm not only resolves the instability issue but also delivers results competitive with the state-of-the-art HSI inpainting algorithms. Unlike previous works where the convergence of the algorithm is either missing or only empirically verified, such as \cite{2021_DIP_In_Loop,lai2022deep, wu2022adaptive}, \cite{pnp_dip}, and \cite{DIP-TV}, {this work has provided theoretical evidence and established convergence guarantees under non-expansiveness and Lipschitz continuity assumptions on the PnP denoiser and the DIP neural network. However, the performance of DIP has been observed to be sensitive to the network structure \cite{sidorov2019deep} and to the training parameters, such as the number of iterations and the learning rate. When the proposed LRS-PnP-DIP(1-Lip) algorithm is given an insufficient number of iterations to learn, DIPs may not work as expected.} In future work, we would like to explore the inductive bias of DHPs, as well as a broader variety of deep learning models, such as transformers ({{the transformer} has recently been explored in the remote sensing community \cite{transformer_remote_sensing_2}}) to exploit their potential in solving HSI inpainting problems. Additionally, the extension of the proposed methods to other HSI processing tasks, such as self-supervised HSI denoising, unmixing, and deblurring, is left to a future investigation.

\vspace{6pt} 

\authorcontributions{{Conceptualization, S.L. and M.Y.; methodology, S.L. and M.Y.; software, S.L.; validation, S.L. and M.Y.; formal analysis, S.L. and M.Y.; investigation, S.L. and M.Y.; resources, S.L.; data curation, S.L.; writing—original draft preparation, S.L; writing—review and editing, S.L. and M.Y.; visualization, S.L. and M.Y.; supervision, M.Y. All authors have read and agreed to the published version of the manuscript.}}
\funding{{This research received no external funding.}}
\dataavailability{{The code for reproducing results is available at \texorpdfstring{\url{https://github.com/shuoli0708/LRS-PnP-DIP}}{\text{https://github.com/shuoli0708/LRS-PnP-DIP}}, accessed on 14 January 2025. The data presented in this study are available upon request from the corresponding author.} }
\conflictsofinterest{The authors declare no conflicts of interest.}


\appendixstart
\appendix
\section[\appendixname~\thesection]{\label{app}}
\appendixtitles{yes}

\subsection[\appendixname~\thesubsection]{On the Design of 1-Lipschitz DIP}\label{appendix_1_Lip_DIP}

{In the} 
 proposed LRS-PnP-DIP(1-Lip) algorithm, we redesigned the DIP to have a Lipschitz constant of 1. To do so, we started by taking a deeper look at how the Lipschitz constant is calculated for each layer; then, we showed that the Lipschitz constraint can be enforced to construct a 1-Lipschitz DIP.

\subsubsection[\appendixname~\thesubsubsection]{Convolutional Layer}
The convolutional layer $conv(f)$ works by performing the convolutional operation on the input to extract the features. The output of the convolution operation can be \mbox{expressed as follows}:
\begin{equation}  \label{convolutional_layer}
\begin{aligned}
     conv(f) = \mathrm{W} \cdot f
\end{aligned}
\end{equation}
where $\mathrm{W}$ is a double-block circular matrix. Substituting \eqref{convolutional_layer} into the definition of the Lipschitz constant, we have the following:
\vspace{-4pt}
\begin{equation} \label{Plug_In_W}
\begin{aligned}
     &\Vert(\mathrm{W} \boldsymbol{x}_1- \mathrm{W}\boldsymbol{x}_2 )\Vert^2 \le L \Vert(\boldsymbol{x}_1- \boldsymbol{x}_2)\Vert^2 \\
     &\Leftrightarrow \Vert \mathrm{W} (\boldsymbol{x}_1- \boldsymbol{x}_2)\Vert^2 \le L \Vert(\boldsymbol{x}_1- \boldsymbol{x}_2)\Vert^2 \\
    &\Leftrightarrow \Vert \mathrm{W} x\Vert^2 \le L \Vert (x)\Vert^2 \quad\text{(Let $\boldsymbol{x} =\boldsymbol{x}_1- \boldsymbol{x}_2$, $\boldsymbol{x}_1 \neq \boldsymbol{x}_2$)}\\
\end{aligned}
\end{equation}\par
{It} follows that\\
\begin{equation} \label{L_upper_bounded_by_W}
\begin{split}
     & \quad \quad \quad \quad \quad \quad \frac{\Vert \mathrm{W} (x)\Vert^2}{\Vert x\Vert^2} \le L \\
    & \text{or equivalently,} \quad L_{\textit{smallest}} = \underset{x\ne 0}{\textit{sup}} \frac{\Vert \mathrm{W} x\Vert^2}{\Vert x\Vert^2} = \Vert \mathrm{W}\Vert^2 
\end{split}
\end{equation} \par
{The} above expression reveals the fact that the spectral norm of $\mathrm{W}$ is bounded by the Lipschitz constant $L$. Hence, regularizing the convolutional layer $conv(f)$ to have Lipschitz continuity boils down to constraining the matrix $\mathrm{W}$; e.g., in our case, $\mathrm{W}$ is expected to have all of its singular values be at most 1, meaning that $conv(f)$ is 1-Lipschitz. To design such a matrix, $\mathrm{W}$, we adopted the power iteration method, as suggested in \cite{henry_gouk_2021regularisation}. The power iteration method works iteratively and approximates the largest singular value of $\mathrm{W}$. Given a sufficient number of iterations, it converges to the true spectral norm of $\mathrm{W}$ with almost negligible errors. In practice, this can be achieved with only a few iterations, which is quite efficient compared to the SVD method.

\subsubsection[\appendixname~\thesubsubsection]{Skip Connection Layer}\label{section_skip_connection}
The DIP has skip connections between the encoder and the decoder. These skip/residual connections ensure anuninterrupted gradient flow, which can effectively tackle the vanishing gradient problem. Moreover, it allows some previous features to be reused later in the network. In particular, for our HSI inpainting problem, we expect some information about the clean pixels from the background to be carried to the next layers.\\ The skip connections take the following form:
\begin{equation} \label{skip_connection}
\begin{split}
     skip(f) = f +  \mathrm{W}(f)
\end{split}
\end{equation}
where $\mathrm{W}$ stands for the composition of possible different functions or layers (e.g., fully connected layers and activation functions). Since the skip connections are additive operators, the Lipschitz constant is determined by both of its components. Suppose that functions $f_1$ and $f_2$, are Lipschitz continuous with the Lipschitz constants $L_1$ and $L_2$, respectively. Then, the composition $f^{'} = f_1 + f_2$ satisfies the following: \\
\begin{equation} \label{skip_connection_lipschitz}
\begin{aligned}    
   \Vert f^{'}(\boldsymbol{x}_1-\boldsymbol{x}_2)\Vert &= \Vert (f_1(\boldsymbol{x}_1)+f_2(\boldsymbol{x}_1))-  (f_1(\boldsymbol{x}_2)+f_2(\boldsymbol{x}_2)) \Vert^2 \\     
     &= \Vert (f_1(\boldsymbol{x}_1)-f_1(\boldsymbol{x}_2)) +(f_2(\boldsymbol{x}_1 -f_2(\boldsymbol{x}_2)) \Vert^2
\end{aligned}
\end{equation}\par
{Applying} the triangular inequality to the $L_2$ norm, we have the following:\\
\begin{equation} \label{skip_connection_triangular}
\begin{aligned}
   \Vert (f_1(\boldsymbol{x}_1)&-f_1(\boldsymbol{x}_2)) +(f_2(\boldsymbol{x}_1 -f_2(\boldsymbol{x}_2)) \Vert^2 \\
   &\le  \Vert f_1(\boldsymbol{x}_1)-f_1(\boldsymbol{x}_2)\Vert^2 
   + \Vert f_2(\boldsymbol{x}_1) -f_2(\boldsymbol{x}_2) \Vert^2
   \end{aligned}
\end{equation}\par
{Furthermore}, the right-hand side of inequality \eqref{skip_connection_triangular} can be bounded by the Lipschitz constant of functions $f_1$ and $f_2$ as follows:
\begin{equation} \label{L1_L2_applied}
\begin{aligned}
 \Vert f_1(\boldsymbol{x}_1)- & f_1(\boldsymbol{x}_2)\Vert^2 + \Vert f_2(\boldsymbol{x}_1) -f_2(\boldsymbol{x}_2) \Vert^2 \\
 &\le L_1 \Vert\boldsymbol{x}_1-\boldsymbol{x}_2 \Vert^2 + L_2 \Vert\boldsymbol{x}_1-\boldsymbol{x}_2 \Vert^2 \\
 &= (L_1+ L_2) \Vert\boldsymbol{x}_1-\boldsymbol{x}_2 \Vert^2
   \end{aligned}
\end{equation}\par
{We} can conclude that the Lipschitz constant of $f^{'} = f_1 + f_2$ is bounded by the sum of the Lipschitz constants of $f_1$ and $f_2$. Therefore, for the skip connections defined in \eqref{skip_connection}, the Lipschitz constant is bounded by $1+L_{\mathrm{W}}$, with a non-negative $L_{\mathrm{W}}$. Because the skip connections would have a Lipschitz constant exceeding 1, we removed them from the implementation of the 1-Lipschitz DIP.

\subsubsection[\appendixname~\thesubsubsection]{Pooling Layer}
The pooling layer $pool(f)$ works by dividing the input image into a set of sub-image patches and converting them to a single value. We used max-pooling in the encoder to extract feature maps from the input HSIs. Max pooling takes the maximum values of the neurons and discards the rest: \\
\begin{equation} \label{max_pooling}
\begin{split}
     pool(f) = max(f_1,f_2,f_3....f_n)
\end{split}
\end{equation}
where n denotes the number of image patches. The expression above can be seen as an affine transformation with a fixed weight; i.e.,
\begin{equation} \label{max_pooling_affine}
\begin{split}
    & \quad \quad \quad  pool(f) = \mathrm{W} \cdot f \\
    & \mathrm{W}_i = \begin{cases}
                1   &  \text{if} \quad f_i = max(f_1,f_2,f_3....f_n)\\
                0        & \text{otherwise}
\end{cases}
\end{split}
\end{equation}\par
{The} gradient of $\mathrm{W}$ is 1 for the neurons with the maximum value and 0 otherwise; the latter does not participate in the propagation. Hence, by definition, the max pooling layer is Lipschitz-continuous with a Lipschitz constant of 1.

\subsubsection[\appendixname~\thesubsubsection]{Activation Layer}
The Lipschitz constant of the activation layer $act(f)$ is trivial, as most activation functions such as ReLu, Leaky-ReLu, Tanh, and Sigmoid have Lipschitz constants of 1.

\subsubsection[\appendixname~\thesubsubsection]{Batch Normalization Layer}
The batch normalization layer $BN(f)$ works by scaling and normalizing the inputs so as to mitigate issues related to internal covariance shifts. It is defined as follows: \\
\begin{equation} \label{BN_layer}
\begin{aligned}    
     BN(f) = \gamma \cdot \frac{f-\mathbb{E}(f)}{\sqrt{\textit{Var}(f)}} + b
\end{aligned}    
\end{equation}\par
{In} the above formulation, $\mathbb{E}(f)$ is the mini-batch mean, and $\textit{Var}(f)$ is the mini-batch variance. $f$ is normalized, scaled by $\gamma$, and shifted with respect to the bias $b$. $\gamma$ and $b$ are two learned parameters during training, which forces the network to also learn the identity transformation and the ability to make a decision on how to balance them. From \eqref{BN_layer}, it is evident that $BN(f)$ is an affine transform with the transformation matrix $\rm W$ being as follows:
\begin{equation}
\begin{aligned}    
      \mathrm{W} = \text{diag}\left (\frac{\gamma}{\sqrt{\textit{Var}(f)}}  \right)
\end{aligned}    
\end{equation}\par
{Therefore}, similar to the analysis of the convolutional layers, the Lipschitz constant of the batch normalization layer is the spectral norm of $\mathrm{W}$. However, $BN(f)$ typically normalizes its input by the variance and scales it by the factor $\gamma$; this would inevitably destroy the Lipschitz continuity. Hence, we propose a variant to the batch normalization layer as follows: \\
\begin{equation} \label{Reduced_BN_Layer}
\begin{aligned}    
     BN^{'}(f) = L \cdot (f-\mathbb{E}(f)) + b 
\end{aligned}    
\end{equation}\par
{The} modified BN operator $BN^{'}$ centers its input by subtracting the mean, which is quite different from its original form, \eqref{BN_layer}. In the experiments, we observed that such an adaptation did not affect the performance of the DIP.

\subsubsection[\appendixname~\thesubsubsection]{Lipschitz Constant of the Full Network}
Consider a neural network, $f_{net}$, which is a composition of $n$ independent sub-networks, $f_1,f_2,...f_n$, each of which has its own Lipschitz constant, $L_1,L_2,...L_n$: \\
\begin{equation} \label{Lipschitz_composition}
\begin{aligned}    
    f_{net}(x) = (f_1 \circ f_2 \circ ... f_n)(x)
\end{aligned}    
\end{equation}\par
{Then}, the Lipschitz constant of $f_{net}$ can be upper-bounded by the following: \\
\begin{equation} \label{Lipschitz_net_upper_bound}
\begin{aligned}    
   L_{net} &\le L_1 \cdot  L_2 \cdot ... \cdot  L_n \\
            &= \prod_{i=1}^{n}L_i
\end{aligned}    
\end{equation}\par
{Hence}, we can tackle the Lipschitz constant of each layer independently and combine them to construct the upper bound of the entire network. We note that the bound certified in \eqref{Lipschitz_net_upper_bound} is not tight and that there are spaces for establishing a tighter bound by considering the network as a whole, which we will leave for future work.
\subsubsection[\appendixname~\thesubsubsection]{Enforcing Lipschitz constraint}\par
{Inspired} by the work in \cite{henry_gouk_2021regularisation}, we introduce an extra projection step after each weight update as follows: \\
\begin{equation} \label{Projection_W}
\begin{aligned}    
     P(\mathrm{W},L) = \frac{1}{\text{max}(1,\Vert \mathrm{W} \Vert^2 / L)} \mathrm{W}
\end{aligned}    
\end{equation}
where $L$ stands for the desired Lipschitz constant, particularly $L=1$ for 1-Lipschitz DIP. This step aims to project the weight matrix, $W$, back to the feasible set for those layers that violate the Lipschitz constraint.

\subsection[\appendixname~\thesubsection]{Proof of Theorem 1}\label{proof_convergence}
The proof relies on the Lyapunov stability theory, which is the heart of the dynamic system analysis \cite{shevitz1994lyapunov}. Lyapunov stability theory can be categorized into (a) the indirect method, which analyzes the convergence through the system state equation, and (b)the direct method, which explicitly describes the behavior of the system's trajectories and its convergence by making use of the Lyapunov function. We refer the reader to \cite{bof2018lyapunov} for a more detailed definition of the Lyapunov function (specifically, Theorem 1.2 and \mbox{Theorem 3.3}). 
In some contexts, it is also known as an energy or dissipative function \cite{hill1976stability}. Compared to the former, the direct method is more appealing, as the convergence can be established by only showing the existence of such a function. In this proof, we begin by defining a function for our proposed LRS-PnP-DIP(1-Lip) algorithm, and we prove its validity. {That is, we construct an energy function, $H^k$, that describes the behaviors of each states in \mbox{Algorithm 3}, and to prove that $H^k$ is a non-increasing function as iteration proceeds; hence, the total energy of the dynamic system established via Algorithm 3 is conservative and stable.} \\
Let $H^k = 2\Vert \boldsymbol{x}^k-\boldsymbol{x}^*\Vert^2 +\frac{1} {\boldsymbol{\mu}^2}\Vert \boldsymbol{\lambda}_1^k-\boldsymbol{\lambda}_1^*\Vert^2 +\frac{1} {\boldsymbol{\mu}^2}\Vert \boldsymbol{\lambda}_2^k-\boldsymbol{\lambda}_2^*\Vert^2$, for a non-zero $\boldsymbol{\mu}$. \\
\textit{Remarks}. This design follows similar structures as in the original convergence proof of the ADMM algorithm \cite{ADMM} and in a recent work \cite{zhang2019fundamental_ADMM}. Here, $H^k$ is a function of the system's state change, which is, by design, non-negative. The first two assumptions in \mbox{Theorem 1.2 \cite{bof2018lyapunov}} automatically hold. Thus, we only need to show that the proposed candidate $H^k$ is a non-increasing function in order to be a valid Lyapunov function.
More specifically, we will show that $H^k$ is a
decreasing function, which satisfies the following: \\
\begin{equation}
\begin{aligned}
  H^k-H^{k+1} \ge C
\end{aligned}
\end{equation}
where C is a positive constant. \par
{Firstly}, recall that the LRS-PnP-DIP(1-Lip) algorithm takes the following update steps:
\begin{equation}  \label{alpha_k+1}
\begin{aligned}
 {\boldsymbol{\alpha}^{k+1}}=\mathcal{T}(\boldsymbol{x}^{k} + \frac{\boldsymbol{\lambda}_1^k}{\boldsymbol{\mu}})
\end{aligned}
\end{equation}

\begin{equation}
\begin{aligned}
  {\boldsymbol{u}^{k+1}}=f_\theta(\boldsymbol{x}^{k}+\frac{\boldsymbol{\lambda}_2^k}{\boldsymbol{\mu}})
\end{aligned}
\end{equation}
\begin{equation} \label{x_k+1}
\begin{aligned}
  {\boldsymbol{x}^{k+1}} = \argmin_{\boldsymbol{x}} \Vert \boldsymbol{y} -\mathrm{M}\boldsymbol{x} \Vert_{2}^2 + \frac{\boldsymbol{\mu}}{2}\Vert(\boldsymbol{x} + \frac{\boldsymbol{\lambda}_1^k}{\boldsymbol{\mu}})-\Phi\boldsymbol{\alpha}^{k+1} \Vert_{2}^2 \\
  + \frac{\boldsymbol{\mu}}{2}\Vert (\boldsymbol{x}+\frac{\boldsymbol{\lambda}_2^k}{\boldsymbol{\mu}}) - \boldsymbol{u}^{k+1} \Vert_{2}^2 \\
\end{aligned}
\end{equation}
\begin{equation} \label{lagrangian}
\begin{aligned}
  {\boldsymbol{\lambda}_1^{k+1}} = \boldsymbol{\lambda}_1^{k} + \boldsymbol{\mu}(\boldsymbol{x}^{k+1}-\Phi \boldsymbol{\alpha}^{k+1})\\
   {\boldsymbol{\lambda}_2^{k+1}} = \boldsymbol{\lambda}_2^{k} + \boldsymbol{\mu}(\boldsymbol{x}^{k+1}-\boldsymbol{u}^{k+1})
\end{aligned}
\end{equation}\par
{We} define $\boldsymbol{x}_e^{k}=\boldsymbol{x}^{k}-\boldsymbol{x}^{*}$, $\boldsymbol{u}_e^{k}=\boldsymbol{u}^{k}-\boldsymbol{u}^{*}$, $\boldsymbol{\alpha_e}^{k}=\boldsymbol{\alpha}^{k}-\boldsymbol{\alpha}^{*}$, $\boldsymbol{\lambda}_{1e}^{k}=\boldsymbol{\lambda}_1^{k}-\boldsymbol{\lambda}_1^{*}$, and $\boldsymbol{\lambda}_{2e}^{k}=\boldsymbol{\lambda}_2^{k}-\boldsymbol{\lambda}_2^{*}$ in the subsequent proof for simplicity. This results in the following: \\
\begin{equation} \label{define_for_simplicity}
\left\{
\begin{aligned}
\boldsymbol{x}_e^{k+1}=\boldsymbol{x}^{k+1}-\boldsymbol{x}^{*} \\
\boldsymbol{u}_e^{k+1}=\boldsymbol{u}^{k+1}-\boldsymbol{u}^{*} \\
\boldsymbol{\alpha}_e^{k+1}=\boldsymbol{\alpha}^{k+1}-\boldsymbol{\alpha}^{*} \\
\boldsymbol{\lambda}_{1e}^{k+1}=\boldsymbol{\lambda}_1^{k+1}-\boldsymbol{\lambda}_1^{*} \\
\boldsymbol{\lambda}_{2e}^{k+1}=\boldsymbol{\lambda}_2^{k+1}-\boldsymbol{\lambda}_2^{*}
\end{aligned}
\right.
\end{equation} \par
{We} take Equation \eqref{x_k+1} as our starting point and denote the first term, $ \Vert \boldsymbol{y} -\mathrm{M}\boldsymbol{x} \Vert_{2}^2$, as $f(\boldsymbol{x})$. The first-order optimality of equation \eqref{x_k+1} implies the following: \\
\begin{equation}
\begin{aligned} \label{1_order_op}
 \nabla f\boldsymbol{(x)} &+\boldsymbol{\mu}(\boldsymbol{x}+\frac{\boldsymbol{\lambda}_1^k}{\boldsymbol{\mu}}-\Phi \boldsymbol{\alpha}^{k+1}) \\
 &+\boldsymbol{\mu}(\boldsymbol{x}+\frac{\boldsymbol{\lambda}_2^k}{\boldsymbol{\mu}}-\boldsymbol{u}^{k+1}) =0
\end{aligned}
\end{equation}\par
{Since} the minimizer $\boldsymbol{x}^{k+1}$ satisfies \eqref{1_order_op}, we plug it into \eqref{1_order_op} to obtain the following: \\
\begin{equation}
\begin{aligned} \label{delta_x_k+1}
 \nabla f\boldsymbol{(x)}^{k+1} +\boldsymbol{\mu}(\boldsymbol{x}^{k+1}-\Phi \boldsymbol{\alpha}^{k+1}+\frac{\boldsymbol{\lambda}_1^k}{\boldsymbol{\mu}}) \\
 +\boldsymbol{\mu}(\boldsymbol{x}^{k+1}-\boldsymbol{u}^{k+1}+\frac{\boldsymbol{\lambda}_2^k}{\boldsymbol{\mu}})  \\
 \overset{\eqref{lagrangian}}{=}\nabla f\boldsymbol{(x)}^{k+1} +\boldsymbol{\mu}(\frac{\boldsymbol{\lambda}_1^{k+1}-\boldsymbol{\lambda}_1^k}{\boldsymbol{\mu}}+\frac{\boldsymbol{\lambda}_1^k}{\boldsymbol{\mu}}) \\
 +\boldsymbol{\mu}(\frac{\boldsymbol{\lambda}_2^{k+1}-\boldsymbol{\lambda}_1^k}{\boldsymbol{\mu}}+\frac{\boldsymbol{\lambda}_2^k}{\boldsymbol{\mu}}) \\
 =\nabla f\boldsymbol{(x)}^{k+1}+\boldsymbol{\lambda}_1^{k+1}+\boldsymbol{\lambda}_2^{k+1} =0
\end{aligned}
\end{equation}\par
{The} critical point satisfies, i.e., $k\rightarrow \infty$: \\
\begin{equation}
\begin{aligned}\label{delta_x_*}
\nabla f\boldsymbol{(x)}^{*}+\boldsymbol{\lambda}_1^{*}+\boldsymbol{\lambda}_2^{*} =0
\end{aligned}
\end{equation}\par
{Due} to the strong convexity of $f(\boldsymbol{x})$, using Lemma \ref{Lemma 2} with $\boldsymbol{x}=\boldsymbol{x}^{k+1}$ and $\boldsymbol{y}=\boldsymbol{x}^{*}$ yields the following: \\
\begin{equation}
\begin{aligned}
   \left \langle \nabla f\boldsymbol{(x)}^{k+1}-\nabla f\boldsymbol{(x)}^{*}, \boldsymbol{x}^{k+1}-\boldsymbol{x}^{*}\right \rangle  &\ge  \rho \Vert(\boldsymbol{x}^{k+1}-\boldsymbol{x}^{*})\Vert^2 
\end{aligned}
\end{equation} \par
{Combining} the above inequality with Equations \eqref{delta_x_k+1} and \eqref{delta_x_*}, we have the following: \\
\begin{equation} \label{lambda_inequality}
\begin{aligned}
   \left \langle -\boldsymbol{\lambda}_{1e}^{k+1}-\boldsymbol{\lambda}_{2e}^{k+1}, \boldsymbol{x}_e^{k+1} \right \rangle &\ge  \rho \Vert\boldsymbol{x}_e^{k+1}\Vert^2 
\end{aligned}
\end{equation}\par
{Secondly}, using Assumption \ref{Assump 2}, the DHP $f_\theta(z)$ is L-Lipschitz with $L\le1$ and $\boldsymbol{x}=\boldsymbol{x}^{k}+\frac{\boldsymbol{\lambda}_2^k}{\boldsymbol{\mu}}$, $\boldsymbol{y}=\boldsymbol{x}^{*}+\frac{\boldsymbol{\lambda}_2^*}{\boldsymbol{\mu}}$; we get the following:\\
\begin{equation}\label{u_inequality}
\begin{aligned}
     \Vert\boldsymbol{u}^{k+1}-\boldsymbol{u}^{*}\Vert^2 \le  \Vert(\boldsymbol{x}^{k}-\boldsymbol{x}^{*})+ \frac{\boldsymbol{\lambda}_2^{k}-\boldsymbol{\lambda}_2^*}{\boldsymbol{\mu}}\Vert^2 \\
     or \Vert \boldsymbol{u}_{e}^{k+1}\Vert^2 \le \Vert \boldsymbol{x}_e^{k}+ \frac{\boldsymbol{\lambda}_{2e}^{k}}{\boldsymbol{\mu}}\Vert^2 \\
\end{aligned}
\end{equation}\par
{Thirdly}, using Assumption \ref{Assump 1} and the resulting Lemma \ref{Lemma 1}, suggesting that the operator $\mathcal{T}$ used in the $\boldsymbol{\alpha}$ updating step \eqref{alpha_k+1} is $\theta$-averaged, and  $\boldsymbol{x}=\boldsymbol{x}^{k}+\frac{\boldsymbol{\lambda}_1^k}{\boldsymbol{\mu}}$, $\boldsymbol{y}=\boldsymbol{x}^{*}+\frac{\boldsymbol{\lambda}_1^*}{\boldsymbol{\mu}}$, we have the following:
\begin{equation}\label{alpha_inequality}
\begin{aligned}
     \Vert\boldsymbol{\alpha}^{k+1}-\boldsymbol{\alpha}^{*}\Vert^2 \le  \Vert(\boldsymbol{x}^{k}-\boldsymbol{x}^{*})+ \frac{\boldsymbol{\lambda}_1^{k}-\boldsymbol{\lambda}_1^*}{\boldsymbol{\mu}}\Vert^2 \\
     or \Vert \boldsymbol{\alpha}_{e}^{k+1}\Vert^2 \le \Vert \boldsymbol{x}_e^{k}+ \frac{\boldsymbol{\lambda}_{1e}^{k}}{\boldsymbol{\mu}}\Vert^2 \\
\end{aligned}
\end{equation} \par
{Now}, we multiply \eqref{lambda_inequality} on both sides by $\frac{2}{\boldsymbol{\mu}}$, and we gather the resulting inequality with \eqref{u_inequality} and \eqref{alpha_inequality}:
\begin{equation}\label{gather_three}
\left\{
\begin{aligned}
   -\frac{2}{\boldsymbol{\mu}}\left \langle \boldsymbol{\lambda}_{1e}^{k+1}, \boldsymbol{x}_e^{k+1} \right \rangle - \frac{2}{\boldsymbol{\mu}}\left \langle \boldsymbol{\lambda}_{2e}^{k+1}, \boldsymbol{x}_e^{k+1} \right \rangle &\ge  \frac{2}{\boldsymbol{\mu}}\rho \Vert\boldsymbol{x}_e^{k+1}\Vert^2 \\
   \Vert \boldsymbol{x}_e^{k}+ \frac{\boldsymbol{\lambda}_{2e}^{k}}{\boldsymbol{\mu}}\Vert^2 \ge  \Vert \boldsymbol{u}_{e}^{k+1} \Vert^2 \\
   \Vert \boldsymbol{x}_e^{k}+ \frac{\boldsymbol{\lambda}_{1e}^{k}}{\boldsymbol{\mu}}\Vert^2 \ge \Vert \boldsymbol{\alpha}_{e}^{k+1} \Vert^2 
\end{aligned}
\right.
\end{equation}\par
{We} put $\boldsymbol{\mu}$ inside the left-hand side of the first inequality, and we add them to give the following: 
\begin{equation}
\begin{aligned}
   -2\left \langle \frac{\boldsymbol{\lambda}_{1e}^{k+1}}{{\boldsymbol{\mu}}}, \boldsymbol{x}_e^{k+1} \right \rangle - 2\left \langle \frac{\boldsymbol{\lambda}_{2e}^{k+1}}{{\boldsymbol{\mu}}}, \boldsymbol{x}_e^{k+1} \right \rangle  +
   \Vert \boldsymbol{x}_e^{k}+ \frac{\boldsymbol{\lambda}_{2e}^{k}}{\boldsymbol{\mu}}\Vert^2 \\
   + \Vert \boldsymbol{x}_e^{k} + \frac{\boldsymbol{\lambda}_{1e}^{k}}{\boldsymbol{\mu}}\Vert^2 \ge \frac{2}{\boldsymbol{\mu}}\rho \Vert\boldsymbol{x}_e^{k+1}\Vert^2+\Vert \boldsymbol{u}_{e}^{k+1} \Vert^2+\Vert \boldsymbol{\alpha}_{e}^{k+1} \Vert^2 
\end{aligned}
\end{equation}
which can be written as:
\begin{equation}
\begin{aligned}
   -(\frac{1}{\boldsymbol{\mu}^2}\Vert\boldsymbol{\lambda}_{1e}^{k+1}\Vert^2+ \Vert \boldsymbol{x}_{e}^{k+1}\Vert^2- \Vert \frac{\boldsymbol{\lambda}_{1e}^{k+1}}{{\boldsymbol{\mu}}}-\boldsymbol{x}_{e}^{k+1}\Vert^2) \\
   -(\frac{1}{\boldsymbol{\mu}^2}\Vert\boldsymbol{\lambda}_{2e}^{k+1}\Vert^2+ \Vert \boldsymbol{x}_{e}^{k+1}\Vert^2- \Vert \frac{\boldsymbol{\lambda}_{2e}^{k+1}}{{\boldsymbol{\mu}}}-\boldsymbol{x}_{e}^{k+1}\Vert^2) \\
   +\frac{1}{\boldsymbol{\mu}^2}\Vert\boldsymbol{\lambda}_{2e}^{k}\Vert^2+ \Vert \boldsymbol{x}_{e}^{k}\Vert^2+ 2\left \langle \frac{\boldsymbol{\lambda}_{2e}^{k}}{\boldsymbol{\mu}},\boldsymbol{x}_{e}^{k}\right \rangle\\
   +\frac{1}{\boldsymbol{\mu}^2}\Vert\boldsymbol{\lambda}_{1e}^{k}\Vert^2+ \Vert \boldsymbol{x}_{e}^{k}\Vert^2+ 
   2\left \langle \frac{\boldsymbol{\lambda}_{1e}^{k}}{\boldsymbol{\mu}},\boldsymbol{x}_{e}^{k}\right \rangle \\
   \ge \frac{2}{\boldsymbol{\mu}}\rho \Vert\boldsymbol{x}_e^{k+1}\Vert^2+\Vert \boldsymbol{u}_{e}^{k+1} \Vert^2+\Vert \boldsymbol{\alpha}_{e}^{k+1} \Vert^2 
\end{aligned}
\end{equation}\par
{After} rearrangement, we get the following:
\begin{equation} \label{recover_H_k}
\begin{aligned}
   2(\Vert \boldsymbol{x}_{e}^{k}\Vert^2 - \Vert \boldsymbol{x}_{e}^{k+1}&\Vert^2)+
   \frac{1}{\boldsymbol{\mu}^2}(\Vert \boldsymbol{\lambda}_{1e}^{k}\Vert^2-\Vert \boldsymbol{\lambda}_{1e}^{k+1}\Vert^2) 
   +\frac{1}{\boldsymbol{\mu}^2}(\Vert \boldsymbol{\lambda}_{2e}^{k}\Vert^2 \\
   &-\Vert \boldsymbol{\lambda}_{2e}^{k+1}\Vert^2) \\
   &\ge \frac{2}{\boldsymbol{\mu}}\rho \Vert\boldsymbol{x}_e^{k+1}\Vert^2+\Vert \boldsymbol{u}_{e}^{k+1} \Vert^2+\Vert \boldsymbol{\alpha}_{e}^{k+1} \Vert^2 \\
   &+\Vert \frac{\boldsymbol{\lambda}_{1e}^{k+1}}{\boldsymbol{\mu}}-\boldsymbol{x}_{e}^{k+1}\Vert^2+
   \Vert \frac{\boldsymbol{\lambda}_{2e}^{k+1}}{\boldsymbol{\mu}}-\boldsymbol{x}_{e}^{k+1}\Vert^2\\
   &-2\left \langle \frac{\boldsymbol{\lambda}_{2e}^{k}}{\boldsymbol{\mu}},\boldsymbol{x}_{e}^{k}\right \rangle -
2\left \langle \frac{\boldsymbol{\lambda}_{1e}^{k}}{\boldsymbol{\mu}},\boldsymbol{x}_{e}^{k}\right \rangle \\
\end{aligned}
\end{equation}\par
{Recall} that\\
\begin{equation} \label{recall_H_k}
\begin{aligned}
H^k& - H^{k+1} \\
&= 2\Vert \boldsymbol{x}^k-\boldsymbol{x}^*\Vert^2 +\frac{1} {\boldsymbol{\mu}^2}\Vert \boldsymbol{\lambda}_1^k-\boldsymbol{\lambda}_1^*\Vert^2 +\frac{1} {\boldsymbol{\mu}^2}\Vert \boldsymbol{\lambda}_2^k -
\boldsymbol{\lambda}_2^*\Vert^2& \\
&-2\Vert \boldsymbol{x}^{k+1}-\boldsymbol{x}^*\Vert^2 -\frac{1} {\boldsymbol{\mu}^2}\Vert \boldsymbol{\lambda}_1^{k+1}-\boldsymbol{\lambda}_1^*\Vert^2 -\frac{1} {\boldsymbol{\mu}^2}\Vert \boldsymbol{\lambda}_2^{k+1}-\boldsymbol{\lambda}_2^*\Vert^2 \\
&\overset{\eqref{define_for_simplicity}}{=}2(\Vert \boldsymbol{x}_{e}^{k}\Vert^2 - \Vert \boldsymbol{x}_{e}^{k+1}\Vert^2) +
   \frac{1}{\boldsymbol{\mu}^2}(\Vert \boldsymbol{\lambda}_{1e}^{k}\Vert^2-\Vert \boldsymbol{\lambda}_{1e}^{k+1}\Vert^2) \\
   &+\frac{1}{\boldsymbol{\mu}^2}(\Vert \boldsymbol{\lambda}_{2e}^{k}\Vert^2-\Vert \boldsymbol{\lambda}_{2e}^{k+1}\Vert^2)
\end{aligned} \\
\end{equation}\par
{It} can be seen that the left-hand side of inequality \eqref{recover_H_k} recovers exactly $H^k-H^{k+1}$. To make $H^k$ a non-increasing function, we require the entire right-hand side to be non-negative; it is thus sufficient to show that the last two terms, $-2\left \langle \frac{\boldsymbol{\lambda}_{2e}^{k}}{\boldsymbol{\mu}},\boldsymbol{x}_{e}^{k}\right \rangle -
2\left \langle \frac{\boldsymbol{\lambda}_{1e}^{k}}{\boldsymbol{\mu}},\boldsymbol{x}_{e}^{k}\right \rangle$, are non-negative. This is straightforward to show if we plug $k=k-1$ into the first line of \eqref{gather_three} to get the following:
\begin{equation}
\begin{aligned}
   H^k-H^{k+1}
   &\ge \frac{2}{\boldsymbol{\mu}}\rho \Vert\boldsymbol{x}_e^{k+1}\Vert^2+\Vert \boldsymbol{u}_{e}^{k+1} \Vert^2+\Vert \boldsymbol{\alpha}_{e}^{k+1} \Vert^2 \\
   &+\Vert \frac{\boldsymbol{\lambda}_{1e}^{k+1}}{\boldsymbol{\mu}}-\boldsymbol{x}_{e}^{k+1}\Vert^2+ 
   \Vert \frac{\boldsymbol{\lambda}_{2e}^{k+1}}{\boldsymbol{\mu}}-\boldsymbol{x}_{e}^{k+1}\Vert^2\\
   &+\frac{2}{\boldsymbol{\mu}}\rho \Vert\boldsymbol{x}_e^{k}\Vert^2\\
   &\ge 0
\end{aligned}
\end{equation}\par
{Now}, if we add both sides of \eqref{recall_H_k}, from $k=0$ to $k=\infty$, it follows that
\begin{equation}
\begin{aligned}
  \sum\limits_{k=0}^{\infty} \frac{2}{\boldsymbol{\mu}}\rho \Vert\boldsymbol{x}_e^{k+1}\Vert^2+ \sum\limits_{k=0}^{\infty}\Vert \boldsymbol{u}_{e}^{k+1} \Vert^2 &+ \sum\limits_{k=0}^{\infty}\Vert \boldsymbol{\alpha}_{e}^{k+1} \Vert^2  \\
  &\le H^0-H^{\infty} < \infty
\end{aligned}
\end{equation}\par
{We} can conclude that sequences ${\boldsymbol{x}_e^{k+1}}$, {$\boldsymbol{u}_e^{k+1}$} and {$\boldsymbol{\alpha}_e^{k+1}$} are all bounded sequences due to the Lyapunov theorem. That is, as $k\rightarrow \infty$: \\
\begin{equation}
\begin{aligned}
  \lim_{k\to\infty} \Vert\boldsymbol{x}^{k+1}-\boldsymbol{x}^{*}\Vert^2 \rightarrow 0 \\
  \lim_{k\to\infty} \Vert\boldsymbol{u}^{k+1}-\boldsymbol{u}^{*}\Vert^2 \rightarrow 0 \\
  \lim_{k\to\infty} \Vert\boldsymbol{\alpha}^{k+1}-\boldsymbol{\alpha}^{*}\Vert^2 \rightarrow 0 
\end{aligned}
\end{equation}\par
{Thus}, the iterations generated via the LRS-PnP-DIP(1-Lip) algorithm converge to the critical points ($\boldsymbol{x}^{*},\boldsymbol{u}^{*},\boldsymbol{\alpha}^{*}$) with a sufficiently large k, and all the trajectories are bounded. LRS-PnP-DIP(1-Lip) is also asymptotically stable according to Theorem 3.3 \cite{bof2018lyapunov}.

\subsection[\appendixname~\thesubsection]{Parameters Tuning and Ablation Tests} \label{Appendix_Parameter Tuning and Ablation Tests}

\subsubsection[\appendixname~\thesubsubsection]{Sensitivity Analysis and Hyperparameters Selections}\label{Ablation_all_hyper_parameters}
{In this section, we provide the detailed selection of hyperparameters for the proposed LRS-PnP-DIP(1-Lip) algorithm in the experiments. In Table \ref{varry_ws_wlr}, we provide the inpainting performance of LRS-PnP-DIP(1-Lip) with the weights of low-rank constraint $w_{lr}$ and sparsity constraints $w_{s}$ varying within the range of 0.1 to 1. It can be seen that the optimal performance is obtained when both low-rank and sparsity constraints are present, as long as there is no overwhelming of either $w_{lr}$ or $w_{s}$. This indicates that LRS-PnP-DIP(1-Lip) is insensitive to the precise value of the low-rank constraint $w_{lr}$ or sparsity constraints $w_{s}$, which is different from the LRS-PnP algorithm \cite{li2023self} discussed in Section \ref{seiton Low-Rank v.s. Sparsity}. Therefore, we set both $w_{lr}$ and $w_{s}$ to 1 across all experiments.}

\begin{table}[H]
\tablesize{\scriptsize}
\caption{{The} 
 selection of hyperparameters for
the proposed LRS-PnP-DIP(1-Lip) algorithm.\label{table_parameter_slection}}
	\begin{adjustwidth}{-\extralength}{0cm}
		\begin{tabularx}{\fulllength}{cccccCCCCC}
			\toprule
\textbf{Parameters} & $\text{\textbf{Number of Iterations}} $ & $\mathbold{\gamma}$  & $\mathbold{w_{lr}}$ & $\mathbold{w_{s}}$ & $\mathbold{\mu_{1}}$ &$\mathbold{\mu_{2}}$ & $\mathbold{\lambda_{1}}$ &$\mathbold{\lambda_{2}}$  &$\mathbold{\sigma_{y}}$ \\
			\midrule
\hspace{1cm}& 200 &1/2 & 1 & 1 & 1/2 & 1/2 & 0 & 0 & 0.12 \\
\midrule
\textbf{Parameters} & $\textbf{DIP Learning Rate}$  & $\textbf{DIP perturbed Noise Level}$ & $\textbf{Lipschitz Constant: L}$  &$\textbf{PnP-ISTA Iterations}$& $\mathbold{\lambda_{ISTA}}$ &\hspace{1cm}&\hspace{1cm}&\hspace{1cm}&\hspace{1cm}\\
\midrule 
\hspace{1cm} & 0.1 & 0 & 1  & 50 & 0.1 &\hspace{1cm}& \hspace{1cm}&\hspace{1cm}&\hspace{1cm}\\ 
			\bottomrule
		\end{tabularx}
	\end{adjustwidth}
\end{table}

%
%


\begin{table}[H]
\caption{{Inpainting} 
 MPSNR of LRS-PnP-DIP(1-Lip) algorithm with varying $w_{lr}$ and $w_{s}$. Using either overwhelming $w_{lr}$ or $w_{s}$ could deteriorate the inpainting results.\label{varry_ws_wlr}}
	\begin{adjustwidth}{-\extralength}{0cm}
		\begin{tabularx}{\fulllength}{CCCCCCC}
			\toprule
\diagbox{$\mathbold{w_{s}}$}{$\mathbold{w_{lr}}$} & \textbf{0} & \textbf{0.1} & \textbf{0.25} &\textbf{ 0.5} & \textbf{0.75} & \textbf{1} \\
			\midrule
\textbf{0} & 30.692 & 37.479($\pm$0.63) & 38.121($\pm$0.46) & 38.475($\pm$0.48) & 38.686($\pm$0.52) & 38.884($\pm$0.31) \\ 
\midrule
\textbf{0.1} & 37.004 & 40.849($\pm$0.31) & 39.107($\pm$0.35) & 39.321($\pm$0.37) & 39.443($\pm$0.30) & 39.812($\pm$0.34)\\ 
\midrule
\textbf{0.25}  & 37.211 & 40.355($\pm$0.29) & 40.798($\pm$0.36) & 40.266($\pm$0.29) & 39.770($\pm$0.23) & 39.906($\pm$0.31)\\ 
\midrule
\textbf{0.5}  & 37.592 & 39.766($\pm$0.27) & 40.807($\pm$0.24) & 40.831($\pm$0.26) & 40.532($\pm$0.25) & 40.245($\pm$0.32)\\ 
\midrule
\textbf{0.75}  & 37.663 & 39.897($\pm$0.37) & 40.373($\pm$0.33) & 40.565($\pm$0.27) & 40.807($\pm$0.30)  & 40.780($\pm$0.26)\\ 
\midrule
\textbf{1}  & 37.957 & 38.591($\pm$0.46) & 39.248($\pm$0.31) & 39.848($\pm$0.25) & 39.902($\pm$0.25) & 40.835($\pm$0.28) \\ 
			\bottomrule
		\end{tabularx}
	\end{adjustwidth}
\end{table}

\subsubsection[\appendixname~\thesubsubsection]{Computational Efficiency}  \label{Computational Efficiency}
{Table \ref{computational_cost} provides per-iteration running time and the total running time upon converge for each method discussed in Section \ref{section_compare_others}. To ensure a fair comparison, all methods, except for GLON \cite{zhao2021tensor}, which uses a pre-trained FFDNet, were implemented with the same DIP backbone. All experiments were conducted in Python with PyTorch 1.13 using the NVIDIA GeForce RTX 3090 GPU. It can be seen that GLON suggests the fastest inference time, but at the cost of sub-optimal inpainting performance. DHP \cite{sidorov2019deep}, \mbox{R-DLRHyIn \cite{niresi2023robust}}, and \mbox{DeepHyIn \cite{DeepHyIn}} share similar computational complexity; these are followed by \mbox{DDS2M \cite{DDS2M}}, which typically requires a large number of diffusion steps in order to yield faithful reconstruction. On the other hand, methods such as DeepRED \cite{2019_Deep_Red}, \mbox{PnP-DIP \cite{pnp_dip}} and LRS-PnP-DIP(1-Lip) are computationally more expensive, which is mainly due to the use of PnP denoisers as regularizers at each iteration. Although LRS-PnP-DIP(1-Lip) introduces extra computation complexity to solve the sparsity-constrained \mbox{problem \eqref{alpha}}, its overall running time is competitive with that of the DeepRED and PnP-DIP methods, This is mainly because LRS-PnP-DIP(1-Lip) enjoys a stability guarantee, which makes it converge faster with fewer iterations. }

\begin{table}[H] 
\tablesize{\tiny}
\caption{{Different} 
 algorithms and their running times on both the Chikusei and Indian Pines datasets.\label{computational_cost}}
\begin{tabularx}{\textwidth}{CCCCCCCCCC}
\toprule
\textbf{Dataset} & \textbf{Cost} & \textbf{GLON} \cite{zhao2021tensor}& \textbf{DHP} \cite{sidorov2019deep} & \textbf{R-DLRHyIn} \cite{niresi2023robust} & \textbf{DeepHyIn} \cite{DeepHyIn} & \textbf{DDS2M} \cite{DDS2M} & \textbf{DeepRED} \cite{2019_Deep_Red} & \textbf{PnP-DIP} \cite{pnp_dip} & \textbf{LRS-PnP-DIP(1-Lip)} \\
\midrule
\textbf{Chikusei} & \textbf{per-iteration} & 0.138 & 0.083 & 0.089 & 0.102 & 0.114 & 2.412 & 2.542 & 3.622\\ 
\midrule
& \textbf{all iterations} & 69.302 & 89.596 & 106.855 & 95.351 & 382.027 & 482.985 & 473.064 & 394.601\\ 
\midrule 
\textbf{Indian Pines} & \textbf{per-iteration} & 0.104 & 0.109 & 0.115 & 0.136 & 0.149 & 2.732 & 2.967 & 3.905 \\ 
\midrule 
 & \textbf{all iterations} & 74.741 & 97.215 & 115.541 & 103.344 & 401.654 & 494.088 & 474.156 & 425.601\\ %
\bottomrule
\end{tabularx}
\end{table}

%
%

\subsubsection[\appendixname~\thesubsubsection]{Effect of the Lipschitz constraint DIP}\label{Ablation_Lipschitz_DIP}
{As an extension to the LRS-PnP-DIP algorithm, LRS-PnP-DIP(1-Lip) enforces the Lipschitz continuity \eqref{Lipschitz_net_upper_bound} of DIP by applying the Lipschitz constraint to each layer of the neural network independently with Lipschitz constant $L=1$. In the proposed LRS-PnP-DIP(1-Lip) algorithm, we remove the residual connections from the DIP (as was discussed in Appendix \ref{section_skip_connection}). However, changing from Skip-Net to U-Net may potentially lead to reduced performance, which is part of the trade-off with the convergence. In Figure \ref{with_without_Lipschitz}, we provide an ablation test comparing the performance of LRS-PnP-DIP with/without 1-Lipschitz constraint. It can be seen that 1-Lipschitz constraint promotes the stability of the reconstruction process, ensuring the convergence of the algorithm. This improvement comes at the cost of only a slight reduction in inpainting performance.}
\vspace{-8pt}
\begin{figure}[H]
 \includegraphics[width=0.8\textwidth,height=0.68\textwidth]{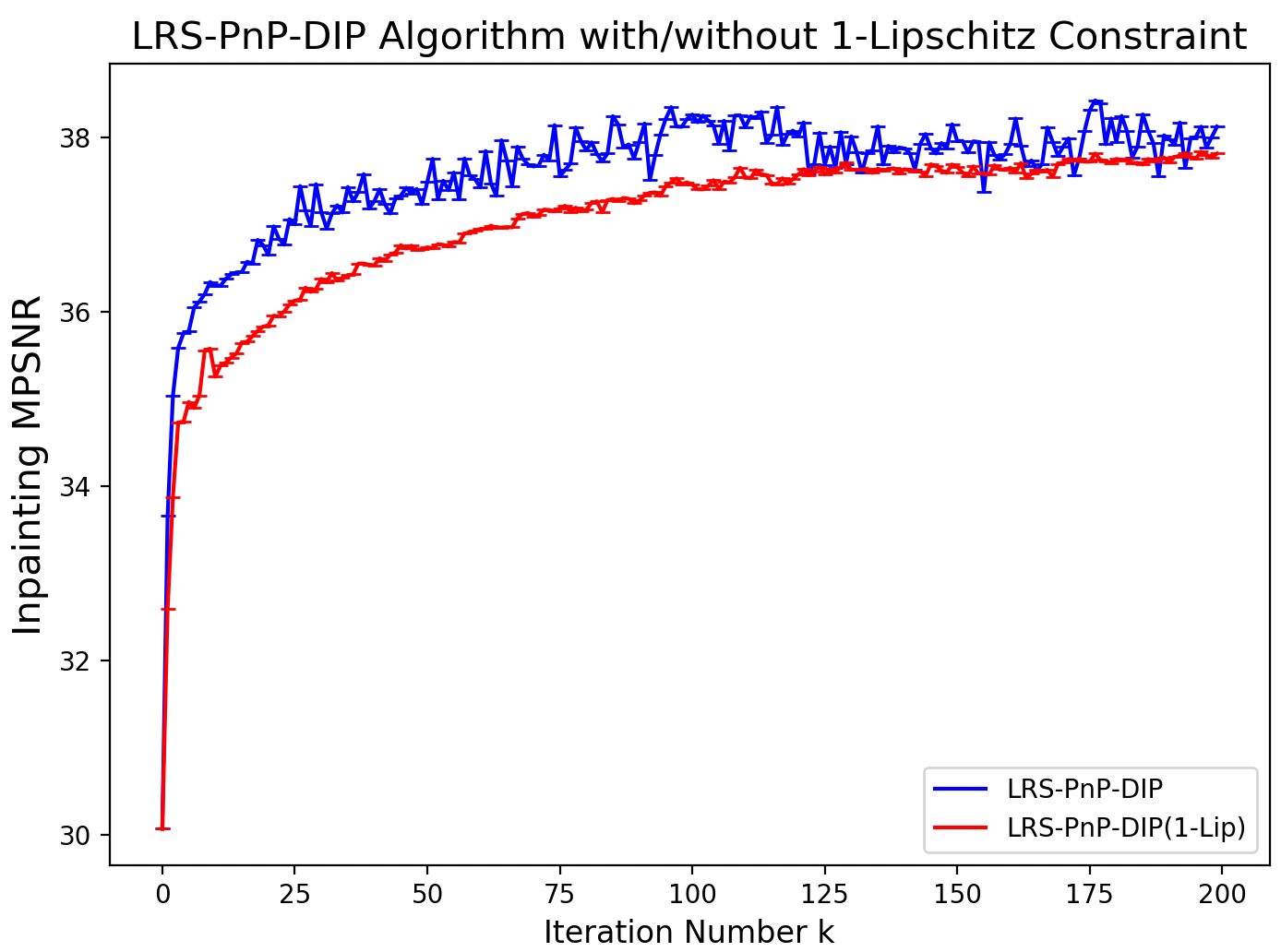}   
 \caption{Inpainting performance of LRS-PnP-DIP algorithm with/without 1-Lipschitz constraint.}
 \label{with_without_Lipschitz}
\end{figure}

\subsubsection[\appendixname~\thesubsubsection]{Effect of the Averaged Denoiser}\label{Ablation_Averaged_Denoiser}
{Compared to the conventional NLM denoiser, the averaged property of the modified NLM denoiser imposes a significantly more restrictive condition. Such a modification also raises the question as to whether it will damage the performance of the original one. In \mbox{Table \ref{averaged_does_not_hurt_performance}}, we record the inpainting MPSNR and MSSIM of the LRS-PnP-DIP algorithm \cite{li2023self} with different PnP denoisers; the LRS-PnP-DIP with averaged NLM denoiser achieves almost the same performance as with the BM3D or the conventional NLM denoiser.}

\begin{table}[H] 
\tablesize{\footnotesize}
\caption{{Inpainting}
 performance of LRS-PnP-DIP algorithm with different PnP denoisers, under noise level $\sigma_y =0.1,0.15,0.2$, respectively. The mean and variance over 20 samples are shown here. {The best results are highlighted in bold.} \label{averaged_does_not_hurt_performance}}
\begin{tabularx}{\textwidth}{crrrrrr}
\toprule
\textbf{Method} & \textbf{Metric} & \textbf{Input} & \textbf{BM3D Denoiser} & \textbf{NLM Denoiser} & \textbf{Averaged NLM {Denoiser} 
} \\
\midrule
$\sigma_y =0.1$ & MPSNR $\uparrow$ &31.76 & 41.23($\pm$0.25) & 41.25($\pm$0.30) & \textbf{41.22}($\pm$0.13) \\ 
\midrule 
\hspace{2cm} & MSSIM $\uparrow$ & 0.304 & 0.920($\pm$0.01) & 0.918($\pm$0.01) & \textbf{0.920}($\pm$0.01)\\ 
\midrule 
$\sigma_y =0.15$ & MPSNR $\uparrow$ &30.38 & 39.55($\pm$0.29) & 39.60($\pm$0.22) & \textbf{39.55}($\pm$0.220)\\ 
\midrule 
\hspace{2cm}  & MSSIM $\uparrow$ & 0.247 & 0.913($\pm$0.01)  & 0.915($\pm$0.01) & \textbf{0.914}($\pm$0.01) \\ 
\midrule 

 $\sigma_y =0.2$ & MPSNR $\uparrow$ &28.75 & 37.68($\pm$0.32) & 37.65($\pm$0.35) & \textbf{37.67}($\pm$0.24)\\ 
\midrule 
\hspace{2cm}  & MSSIM $\uparrow$ & 0.229 & 0.904($\pm$0.01)  & 0.903($\pm$0.01) & \textbf{0.903}($\pm$0.01) \\ 
\bottomrule
\end{tabularx}
\end{table}

\subsubsection[\appendixname~\thesubsubsection]{Effect of the DIP Network Architectures} \label{Ablation_DIP_network_structures}
{As was reported in a series of DIP-related works \cite{ulyanov2018deep,sidorov2019deep,DeepHyIn,pnp_dip}, the performance of DIP is sensitive to the structure of the neural networks. Table \ref{Ablation_Network} provides a comparative analysis of several network architectures, including ResNet 2D/3D, U-Net 2D/3D, and Skip-Net 2D/3D. Empirical results indicate that Skip-Net 2D (i.e., a U-Net architecture with skip connections and 2D convolution layers) achieves the best performance for the HSI inpainting task. Hence, Skip-Net 2D was selected and implemented as the backbone architecture for the proposed LRS-PnP-DIP, LRS-PnP-DIP(1-Lip), and all other competing methods in the comparative analysis. Furthermore, it was found that ResNet architectures yield inferior results within the proposed framework, highlighting the importance of network design choices for achieving optimal inpainting performance. We left the exploration and identification of the most suitable DIP network structure for HSI inpainting as a  future research direction.}

\begin{table}[H] 
\caption{{On}
 the choice of different DIP network architectures in the proposed LRS-PnP-DIP algorithm.\label{Ablation_Network}. {For the metrics MPSNR and MSSIM, higher values indicate better performance. The best results are highlighted in bold.}}
 \centering
\begin{tabular}{@{}ccccc@{}}
\toprule
\textbf{Methods} & \textbf{MPSNR}$\uparrow$ & \textbf{MSSIM}$\uparrow$ \\
\midrule
Input &22.582 & 0.178\\
\midrule 
ResNet 2D & 30.975($\pm$0.62) & 0.610($\pm$0.02)\\
\midrule 
ResNet 3D & 29.661($\pm$1.20) & 0.589($\pm$0.03)\\
\midrule 
UNet 2D  & 35.963($\pm$0.42) & 0.882($\pm$0.02)\\
\midrule 
UNet 3D  & 35.438($\pm$0.56) & 0.868($\pm$0.01)\\
\midrule 
Skip-Net 2D  & $\boldsymbol{37.971}(\pm 0.30)$ & $\boldsymbol{0.899}(\pm 0.01)$\\
\midrule 
Skip-Net 3D  & 37.050($\pm$0.32) & 0.890($\pm$0.01)\\
\bottomrule
\end{tabular}
\end{table}

%
%
%
%

\begin{adjustwidth}{-\extralength}{0cm}

\reftitle{References}

\PublishersNote{}
\end{adjustwidth}
\end{document}